\documentclass[acmsmall]{acmart} %

\usepackage{mathtools}
\usepackage{amsthm}
\usepackage[utf8]{inputenc}

\usepackage[frozencache,cachedir=.]{minted}
\usepackage{csquotes}
\usepackage{balance}
\usepackage{enumitem}
\usepackage{stfloats}
\usepackage{booktabs}
\usepackage{color, colortbl}
\PassOptionsToPackage{hyphens}{url}
\usepackage{adjustbox}
\usepackage[noabbrev, nameinlink]{cleveref}
\usepackage{svg}
\usepackage{meta/circledsteps}
\usepackage{meta/dynamicwidth}

\usepackage{caption}
\usepackage{subfig}

\usepackage{pgfgantt}

\setcopyright{rightsretained}
\acmJournal{PACMMOD}
\acmYear{2025}
\acmVolume{3}
\acmNumber{1 (SIGMOD)}
\acmArticle{55}
\acmMonth{2}
\acmPrice{}
\acmDOI{10.1145/3709705}

\ifluatex
\else
\DeclareUnicodeCharacter{2212}{−}
\fi

\newcommand*{\modyn}{\texorpdfstring{\dynamicwidth{\textsc{Modyn}}{\textsc{Modyn}}}{Modyn}}  %

\definecolor{LightGray}{gray}{0.95}

\definecolor{plotorange}{rgb}{0.8705882352941177, 0.5607843137254902, 0.0196078431372549}
\definecolor{plotred}{rgb}{0.8115340253748559, 0.3211072664359862, 0.2758169934640523}
\definecolor{plotblue}{rgb}{0.2366013071895425, 0.5418685121107266, 0.7470203767781622}
\definecolor{plotpink}{rgb}{0.8, 0.47058823529411764, 0.7372549019607844}

\newcommand{\revision}[1]{#1}
\newcommand{\y}[1]{\cite{#1}}

\definecolor{light_orange}{RGB}{255,184,79}  %

\definecolor{light_red}{RGB}{253,195,183}  %
\definecolor{light_gray}{RGB}{198,230,233}  %
\definecolor{light_blue}{RGB}{230, 230, 255}  %

\definecolor{accent_orange}{RGB}{255, 128, 2}  %
\definecolor{accent_blue}{RGB}{178, 178, 255}  %
\definecolor{accent_blue_dark}{RGB}{102, 102, 255}  %

\definecolor{light_grey}{gray}{0.95}

\NewDocumentCommand{\textganttbar}{ O{fill=light_blue}O{}mmmm }{%
    \ganttbar[bar/.append style={draw=none, fill=none}]{#3}{#5}{#6}
    \ganttbar[
        #2, inline, bar label font=\large, bar/.append style={#1}
    ]{#4}{#5}{#6}
}

\NewDocumentCommand{\textganttbarcur}{ O{}mmmm }{%
    \textganttbar[fill=light_orange][#1]{#2}{#3}{#4}{#5}
}

\NewDocumentCommand{\textganttbarref}{ O{}mmmm }{%
    \textganttbar[fill=light_blue][#1]{#2}{#3}{#4}{#5}
}

\NewDocumentCommand{\textganttbarholdout}{ O{}mmmm }{%
    \textganttbar[fill=white][#1]{#2}{#3}{#4}{#5}
}

\NewDocumentCommand{\textganttbartrain}{ O{}mmmm }{%
    \textganttbar[fill=light_blue][#1]{#2}{#3}{#4}{#5}
}

\NewDocumentCommand{\textganttbareval}{ O{}mmmm }{%
    \textganttbar[fill=light_orange][#1]{#2}{#3}{#4}{#5}
}

\NewDocumentCommand{\textganttmilestone}{ O{}mmm }{%
    \ganttmilestone{#2}{#4}
    \ganttmilestone[#1, inline,bar label font=\large,
        milestone label node/.append style={text background/.style={fill=green}}
    ]{#3}{#4}
}

\NewDocumentCommand{\textganttmilestonewarmup}{ O{}mmm }{%
    \ganttmilestone{#2}{#4}
    \ganttmilestone[#1, inline,bar label font=\large, milestone/.append style={fill=accent_blue, draw=accent_blue_dark, line width=1pt},
        milestone label node/.append style={text background/.style={fill=green}}
    ]{#3}{#4}
}

\NewDocumentCommand{\textganttmilestoneinactive}{ O{}mmm }{%
    \ganttmilestone{#2}{#4}
    \ganttmilestone[#1, inline,bar label font=\large, milestone/.append style={fill=lightgray!30, draw=lightgray!80, line width=1pt},
        milestone label node/.append style={text background/.style={fill=green}}
    ]{#3}{#4}
}

\NewDocumentCommand{\textganttmilestoneactive}{ O{}mmm }{%
    \ganttmilestone{#2}{#4}
    \ganttmilestone[
        #1, inline,bar label font=\large, milestone/.append style={fill=accent_orange},
        milestone label node/.append style={text background/.style={fill=none}}
    ]{#3}{#4}
}

\NewDocumentCommand{\anchorganttmilestone}{ mm }{%
    \ganttmilestone[
        inline,bar label font=\large,
        milestone/.append style={shape=ellipse, draw=red, fill=green},
    ]{#1}{#2}
}

\definecolor{light_grey}{gray}{0.95}
\definecolor{ganttorange}{RGB}{254,213,172}
\definecolor{ganttblue}{RGB}{43,0,249}
\definecolor{ganttdeeporange}{RGB}{253,122,35}

\makeatletter
\gdef\@copyrightpermission{
 \begin{minipage}{0.2\columnwidth}
  \href{https://creativecommons.org/licenses/by/4.0/}{\includegraphics[width=0.90\textwidth]{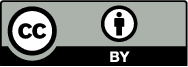}}
 \end{minipage}\hfill
 \begin{minipage}{0.8\columnwidth}
  \href{https://creativecommons.org/licenses/by/4.0/}{This work is licensed under a Creative Commons Attribution International 4.0 License.}
 \end{minipage}
 \vspace{5pt}
}
\makeatother

\begin{document}

\title{\modyn: Data-Centric Machine Learning Pipeline Orchestration}

\author{Maximilian Böther}
\email{mboether@ethz.ch}
\orcid{0000-0003-4093-4361}
\affiliation{%
  \institution{ETH Zurich}
  \country{Switzerland}
}

\author{Ties Robroek}
\email{titr@itu.dk}
\orcid{0009-0006-3451-5602}
\affiliation{%
  \institution{IT University of Copenhagen}
  \country{Denmark}
}

\author{Viktor Gsteiger}
\email{vgsteiger@student.ethz.ch}
\orcid{0000-0002-6750-5500}
\affiliation{%
  \institution{ETH Zurich}
  \country{Switzerland}
}

\author{Robin Holzinger}
\email{robin.holzinger@tum.de}
\orcid{0000-0002-7505-0673}
\affiliation{%
  \institution{Technical University of Munich}
  \country{Germany}
}

\author{Xianzhe Ma}
\email{xianzhema@student.ethz.ch}
\orcid{0009-0000-1400-6735}
\affiliation{%
  \institution{ETH Zurich}
  \country{Switzerland}
}

\author{Pınar Tözün}
\email{pito@itu.dk}
\orcid{0000-0001-6838-4854}
\affiliation{%
  \institution{IT University of Copenhagen}
  \country{Denmark}
}

\author{Ana Klimovic}
\email{aklimovic@ethz.ch}
\orcid{0000-0001-8559-0529}
\affiliation{%
  \institution{ETH Zurich}
  \country{Switzerland}
}

\renewcommand{\shortauthors}{Maximilian Böther et al.}

\begin{abstract}
In real-world machine learning (ML) pipelines, datasets are continuously growing.
Models must incorporate this new training data to improve generalization and adapt to potential distribution shifts.
The cost of model retraining is proportional to how frequently the model is retrained and how much data it is trained on, which makes the naive approach of retraining from scratch each time impractical.

We present \modyn, a data-centric end-to-end machine learning platform. 
\revision{\modyn's ML pipeline abstraction enables users to declaratively describe policies for continuously training a model on a growing dataset.}
\modyn{} pipelines allow users to apply data selection policies (to reduce the number of data points) and triggering policies (to reduce the number of trainings).
\modyn{} executes and orchestrates these continuous ML training pipelines.
The system is open-source and comes with an ecosystem of benchmark datasets, models, and tooling.
\revision{We formally discuss how to measure the performance of ML pipelines by introducing the concept of composite models, enabling fair comparison of pipelines with different data selection and triggering policies.}
We \revision{empirically} analyze how various data selection and triggering policies impact model accuracy, and also show that \modyn{} enables high throughput training with sample-level data selection.

\end{abstract}


\maketitle

\section{Introduction}\label{sec:intro}
The datasets fueling today's production machine learning (ML) models, which typically come from a myriad of sensors or real-time user click streams, are continuously growing~\cite{Tesla2019AutonomyDay, Li2021Shifts,Bhardwaj2022Ekya}.
To maintain high accuracy, stale models deployed in the wild need to be retrained in order to incorporate new data, particularly as training data may experience distribution drifts~\cite{Vela2022Degradation,Schelter2023Argus, Rabanser2019DriftStudy,Liu2023Incremental,Shankar2022MLOpsStudy,Huyen2022RTML,Huyen2022DMLS,He2014Meta,Steidl2023Pipeline,Lazaridou2021MindtheGap,Matam2024Recsysupdate}.
In practice, models may be retrained as often as every day~\cite{Egg2021Grubhub}, while the volume of data that models train on can be as high as petabytes or even exabytes, depending on the application domain~\cite{Zhao2022MetaRM, Maglev2020NVIDIA}.

\begin{figure}
    \centering
    \begin{adjustbox}{trim=0cm 0.05cm 0cm 0.3cm}
        \includesvg[width=0.65\linewidth]{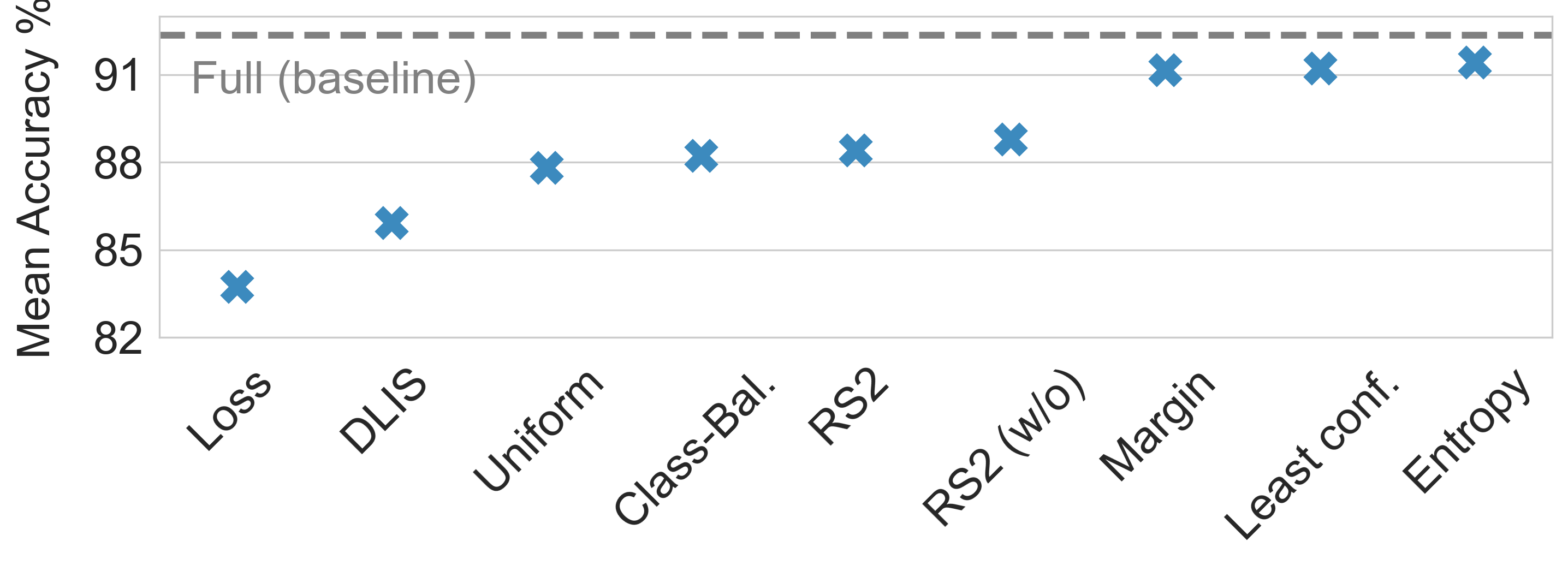}
        \end{adjustbox}
    \caption{\revision{Mean accuracies of 9 selection strategies (50\,\% subset) and full data training (see \Cref{subsubsec:eval-selection-yearbook})}.} 
    \label{fig:motivation-selection}
    \Description{TODO!}
\end{figure}

The cost of continuously training an ML model depends on \textit{how frequently} we retrain the model and \textit{how much data} we use to train the model each time.
The naive approach of retraining a model from scratch on the entire dataset when new data becomes available is prohibitively expensive and slow~\cite{Kasundra2023Framework,Mahadevan2023Retraining}.
To make retraining\footnote{In this paper, retraining refers to both finetuning or training from scratch.} models practical in real use cases, we need to minimize the frequency and the volume of data that a model is trained on, while maintaining high model quality. 
\revision{
For example, \Cref{fig:motivation-selection} shows how various data selection policies (x-axis) proposed in ML literature maintain model accuracy comparable to training on all data (dashed line) while training on only 50\,\% of the \textsc{yearbook} image classification dataset~\cite{Yao2022WildTime}.}
Complementary to data selection, data drift detection can help to trigger retraining only when data characteristics change.
\revision{This can save cost and/or increase model quality compared to fixed-interval retraining schedules.}

However, finding the right data selection and triggering policies is non-trivial.
While ML researchers have explored how to effectively select important samples in a dataset with various strategies~\cite{Paul2021Diet,a,Ramalingam2021Submod,Mirzasoleiman2020CRAIG,Pooladzandi2022AdaCore,Mindermann2022RHOLOSS,Jia2021Shapley,Prabhu2020GDumb,Aljundi2019MIR,Aljundi2019GSS,Kirsch2023DataPruning,b}, it is not clear what policy to use for real-world datasets that grow and exhibit distribution shifts over time. 
ML studies in this space often focus on smaller, static datasets, such as \textsc{Cifar}~\cite{Krizhevsky2009CIFAR} and \textsc{Mnist}~\cite{Lecun1998MNIST}, and do not consider the total pipeline cost, or they focus on one particular metric (e.g., information retention in continual learning studies~\cite{Prabhu2023CGLM, Cai2021CLOC}).  
Several drift detection techniques exist~\cite{Rabanser2019DriftStudy,Tahmasbi2021DriftSurf,Gretton2012TST,Ross2012Drift,Lu2018LearningUnderDrift}. \revision{Existing studies, however, focus on tabular data, synthetically inject drift, and do not train neural networks in response to drift~\cite{Redyuk2021DQV,Schelter2020Validate,Redyuk2024Pipelines,Aguiar2024DriftAnalysis,Barros2018DriftDetectors,Werner2024DriftPerformance}.}
Using such techniques as triggering policies is non-trivial as it involves tuning many hyperparameters.
Most pipelines today are still human-driven~\cite{Huyen2020RTML, Shankar2022MLOpsStudy}.

Furthermore, it is challenging to implement data selection policies in large-scale growing dataset environments while maintaining high training throughput.
Data ingestion is a common bottleneck in ML training~\cite{Murray2021tfdata, Kuchnik2022Plumber, Zhao2022MetaRM}.
Applying data selection policies requires accessing individual data samples rather than sequentially reading input data files. Such random access patterns can degrade training throughput.
In~\Cref{subsec:eval-tput}, we show that multiple levels of batching, parallelizing, and prefetching of reads are essential to achieve high throughput.
Such optimizations should be done transparently by a platform, while ML users focus on defining the logic of ML training and data preprocessing pipelines. 
While others have also acknowledged the need for a continuous training platform that enables users to explore data selection and (re)training policies~\cite{Gantry2021CLSystems, Shankar2022StreamingML, Paleyes2023MLDeployment, Diethe2019CLinPractice}, current \revision{open-source} systems only have limited support for retraining~\cite{Modi2017TFX,Baylor2019TFXCT,Tian2018Continuum,Derakhshan2019Retraining}.
We are not aware of any platform supporting sample-level data selection policies. 

We present \modyn, a data-centric machine learning pipeline orchestrator that addresses this gap.
To the best of our knowledge, \revision{in particular for modalities commonly used in DNN training such as images,} \modyn{} is the first \revision{open-source} orchestrator to support data selection and retraining decisions based on the incoming data.
\modyn{} is an end-to-end platform that supports the entire pipeline lifecycle, including sampling-level data selection and management, triggering model retraining, continuously evaluating model quality, and managing model snapshots.
In this paper, we contribute the following:
\begin{enumerate}[leftmargin=*,nosep]
\item We design an ML pipeline abstraction, which enables users to express how to continuously train a model on growing data.
It allows \revision{declaratively} specifying data-centric policies for model retraining and training data selection, while decoupling the implementation of these policies. 
\revision{We design the abstraction to capture a taxonomy of data selection and triggering policies.}

\item We build \modyn{}, an orchestrator that runs data-centric ML pipelines.
\modyn{} supports various data selection techniques while optimizing for high-throughput sample-level data selection for multiple data formats.
It also supports time-, data volume-, \revision{performance-}, and data drift-based triggering policies while managing and continuously evaluating model versions. 
\modyn{} enables sample-level data selection with comparable throughput to sequentially ingesting data from local storage. 

\item 
\revision{We formalize ML pipelines and introduce \emph{composite models} which describe the performance of a pipeline over its lifetime, and allow for a fair comparison of pipelines with different selection and triggering policies.}
We build an ecosystem around \modyn{} to facilitate policy exploration. 
\modyn{} comes with web-based tooling to compare pipelines in terms of system throughput, training cost, and model quality metrics.
It also comes with a set of benchmark models and datasets with timestamped data for policy evaluation.
For a subset of the accompanying benchmarks, we include case studies on selection and triggering policies, showing how these policies impact pipeline performance. 
\end{enumerate}

\section{Background and Motivation}\label{sec:background}

In this section, we discuss the growing nature of real ML datasets (\Cref{subsec:back-dynamicdata}) and motivate the need for a new platform (\Cref{subsec:back-sys}).

\subsection{Growing Datasets \& ML Perspective}\label{subsec:back-dynamicdata}

Real-world ML datasets are often dynamic, in contrast to static datasets such as \textsc{ImageNet}~\cite{Deng2009ImageNet} that are typically used in ML research~\cite{Bother2023Modyn}.
They either grow as more samples are collected (e.g., from continuous data sources like sensors or click streams) or shrink as data is deleted (e.g., due to privacy reasons).
In this work, we focus on the challenges of training ML models on \textit{growing data}.

\textbf{Why growing data matters.}
Incoming data captures current trends and reveals \emph{distribution shifts} that can be critical in many application domains~\cite{Shankar2022Observability,Vela2022Degradation}, like recommender systems~\cite{Hazelwood2018Meta, He2014Meta,Egg2021Grubhub,Zhao2022MetaRM,Yang2023Updates} and language models~\cite{Lazaridou2021MindtheGap}.
For example, the GrubHub food delivery platform observed a 20\,\% increase in purchase rate when their model is retrained daily rather than weekly~\cite{Egg2021Grubhub}.
Even in the absence of significant distribution shifts, including additional data over time can enhance model performance as it improves generalization.
For example, Tesla continuously collects street pictures to refine their autonomous driving models~\cite{Tesla2019AutonomyDay}.
Growing data impacts training cost, as the cost is proportional to \textit{(i)} how often the model trains and \textit{(ii)} the number of data samples it trains on~\cite{Kasundra2023Framework,Mahadevan2023Retraining}. 

\textbf{ML perspective on growing data.} ML research so far has explored optimizing when to retrain a model and what data to select for training as two isolated dimensions.
The field of \emph{continual learning} (CL)~\cite{Diethe2019CLinPractice, Prabhu2020GDumb, LopezPath2017GEM, Aljundi2019GSS, Aljundi2019MIR, Bang2021RM, Koh2022CLIB}, or incremental learning~\cite{Wu2019LSIL,Polikar2001Learn,Cauwenberghs00SVM}, adapts ML models to ongoing data streams by focusing on learning new tasks, defined as groups of classes.
It is unclear how these techniques apply to real use cases, as CL research has focused on small datasets with synthetic perturbations that lack a true notion of time. 
Furthermore, both the focus on learning classes over time instead of adapting to distribution shift and the common assumption of limited storage are not realistic, as acknowledged by recent works in the CL community~\cite{Prabhu2023CGLM,Ghunaim202Evlauation}.
Data selection policies outside of CL focus on selecting subsets of static datasets~\cite{Mindermann2022RHOLOSS,Mirzasoleiman2020CRAIG,a}.
All techniques require sample-level data access on the dataset.

\revision{While there is work on detecting distribution shift, these papers often focus on theoretical aspects and do not actually train models~\cite{Li2021Shifts,Tahmasbi2021DriftSurf,Rabanser2019DriftStudy,Ross2012Drift, Paz2017TST, Gretton2012TST,Barros2018DriftDetectors,Aguiar2024DriftAnalysis}, i.e., they only compare drift scores.
Notably, Werner et al.~\cite{Werner2024DriftPerformance} train random forests on tabular datasets, and Yuan et al.~\cite{Yuan2022ConceptDrift} consider synthetically perturbed variations of the MNIST dataset from continual learning.
To the best of our knowledge, no paper explored applying drift detection techniques to training large neural networks on modalities such as images and text from non-synthetic benchmarks.
}

\subsection{Platform Support}\label{subsec:back-sys}
Managing when to retrain models and on what data to train models in large-scale growing data environments is challenging. 
It requires efficiently orchestrating continuous training pipelines with configurable triggers and fast access to arbitrary sets of data samples determined by a data selection policy.
Model training orchestration and sample-level data fetching should be transparently optimized by a platform in order to help ML researchers focus on policy exploration and to help ML practitioners reliably deploy ML pipelines in production environments. 
Furthermore, drift detection techniques need to be closely embedded into the data flow, since they typically need to access the previously trained models and the data stream.

Current ML platforms do not address these requirements.
The majority of ML training platforms, such as Weights \& Biases~\cite{Wandb} or MLFlow~\cite{Chen2020MLFlow}, are tailored more towards experiment tracking than continuous retraining.
While a few (often commercial) platforms like NeptuneAI~\cite{NeptuneAI}, Amazon SageMaker~\cite{SageMaker}, Continuum~\cite{Tian2018Continuum}, or Tensorflow TFX~\cite{Modi2017TFX} have partial retraining support, deploying continuous retraining still requires a lot of manual plumbing~\cite{Baylor2019TFXCT,Gantry2021CLSystems,Paleyes2023MLDeployment,Shankar2022MLOpsStudy}.
\revision{Commonly, platforms allow for the performance of the deployed model to trigger a retraining (e.g., SageMaker or tf-serving~\cite{Olston2017tfserving}).
Notably, Hopsworks~\cite{Hopsworks} supports drift detection on individual features of tabular data, and SageMaker's Model Monitor allows for the collection of drift metrics on tabular data using the Deequ library~\cite{Schelter2018Automating}.
Images and other modalities are explicitly not supported currently.}
Platforms such as Ekya~\cite{Bhardwaj2022Ekya}, which optimizes retraining for vision models on edge devices, and  Ekko~\cite{Sima2022Ekko}, which optimizes model updates for recommendation systems, cater to specific use cases.
The aforementioned platforms view the datasets as a big blob of data instead of indexing individual samples.%
Especially for modalities commonly used in DNN training (images and text), data-centric decision making on when and what data to train on is, to the best of our knowledge, not supported by any available open-source training platform.

\section{Modeling Dynamic ML Pipelines}\label{sec:formal-modeling}

\revision{%
Continuous ML pipelines regularly run model trainings on an incoming stream of data \(S\) with a discrete time clock.
The data arrives in batches \(S_i \subset S\), i.e., sequences of new samples, where batch \(t\) is given as \(S_t = (s_1, \ldots, s_{n_t})\).
Each sample \(s_i \in S_t\) comprises a unique identifier, a label, a timestamp, other metadata, and a data payload.
}

\revision{
\textbf{Triggering.}
The triggering policy decides whether to trigger a new training\footnote{We use the terms training and retraining interchangeably.}. 
We model a triggering policy as a function $\pi: \mathcal{P}\left(S\right) \rightarrow \bigcup_{n=0}^\infty \mathcal{P}\left([1, \ldots, n]\right)$.
Given a batch \(S_t\), it determines which samples \(s_i \in S_t\) trigger a new training process.
Formally, it outputs a sequence \( \pi\left( S_t \right) = \left( i \in \left[ 1 \ldots n_t \right] \mid s_i \in S_t \text{ causes trigger} \right)\).
The triggering policy can be stateful and utilize the observed history of samples, properties of them or the pipeline, to come to a triggering decision.
Conceptually, the triggering policy decides on a per-sample basis.
For efficiency, our implementation evaluates multiple samples simultaneously in batches, while keeping the semantics of per-sample decision-making.
Note that triggering on each new data sample is impractical in production, as each newly trained model typically needs to undergo a set of extensive deployment checks, which are expensive to run at high frequency~\cite{Huyen2022DMLS,Shankar2022MLOpsStudy}.
}

\revision{\textbf{Data selection.}
On each trigger, the selection policy chooses which samples to train on.
Let \(s_k \in S_t\) cause the overall \(r\)-th retraining trigger.
The  observed data until trigger \(r\) is \(\mathcal{D}^{\text{tot}}_{r} = \left\{ s_i \in S_t \mid i \leq k \right\} \cup \bigcup_{t' < t} \mathsf{set}\left(S_{t'} \right) \).
A data selection policy is a function $\xi_r: \mathcal{D}^{\text{tot}}_{r} \rightarrow \mathbb{R}^{| \mathcal{D}^{\text{tot}}_r |}$ that assigns each item in the total observed data a weight.
Thereby, the function defines the \(r\)-th \emph{trigger training set}  \(\mathcal{D}_{r} \subset \mathcal{D}^{\text{tot}}_{r} \times \mathbb{R}^+\).
An item is included if its weight is greater than 0.
The data selection policy selects from all previously seen data samples, i.e., they can come from any \(S_{t'}\) with \(t' < t\), and all samples in \(S_t\) until \(s_k\).
The sample weights can be used to prioritize samples by multiplying their gradients with the weights during backpropagation.
The trigger training set is a \emph{subset} of all data points seen so far, so it may, but does not have to, contain samples from previous triggers.
}

\subsection{Evaluating and comparing pipelines}\label{subsec:formal-comparing}

\revision{To compare ML pipelines, we first need to define how to quantify the performance and cost of a pipeline.
Evaluating the model quality and training cost of an ML pipeline on a growing dataset is more complex than evaluating a single model training on a static dataset.}
Two challenges arise.

\revision{First, ML pipelines train multiple models instead of a single one to deal with the growing data that might exhibit distribution shift.
On each retraining trigger \(r\) we train a new model \(m_r\).
For a pipeline \(P\), let \(\mathcal{M}_P\) denote the sequence of all models trained during pipeline execution.
Each \(m_r \in \mathcal{M}_P\) is a 4-tuple containing its model weights \(w_r\), the data it was trained on \(\mathcal{D}_r\), and the start timestamp \(t^s_r\) and end timestamp \(t^e_r\) of the training data, i.e., \(m_r = \left\langle w_r, \mathcal{D}_r, t^s_r, t^e_r \right\rangle\).
We should not just evaluate a single model from \(\mathcal{M}_P\), e.g., the last one, on the entire dataset since the model is trained on one particular distribution.
Instead, we need to consider multiple models.
}

\revision{
Second, to understand how a model's performance changes over time,  we need to define windows over the evaluation data, as discussed by Shankar et al.~\cite{Shankar2022StreamingML}.
Evaluation data should be separate from training data, e.g., by partitioning the stream \(S\).
These windows are temporal slices of the dataset on which we then calculate a  quality metric per model.}
\revision{Let \(P_1\) and \(P_2\) be pipelines with different triggering policies \(\pi_1\) and \(\pi_2\) on the same stream of data. 
\(\mathcal{M}_{P_1}\) and  \(\mathcal{M}_{P_2}\) contain different models, in particular with different timestamps.
The intuitive solution of defining evaluation windows matching the training intervals of the models, i.e., each model \(m_r\) defines an evaluation window from \(t^s_r\) to \(t^e_r\), is not fair across pipelines.
Hence, we first need to decouple determining evaluation intervals from triggering and then define which model to use for which window.}

\revision{
We define an evaluation interval as a 3-tuple \(\left\langle \tau^s, \tau^a, \tau^e \right\rangle\). %
\(\tau^s\) and \(\tau^e\) define the start and end of the range from which we consider evaluation data.
\(\tau^a\) defines the \emph{anchor point} of the interval, which serves as a reference timestamp that we use in further definitions.
Typically, \(\tau^a = \tau^s\) or \(\tau^a = \left( \tau^s + \tau^e\right) / 2\).
We define an \emph{interval generation function} \(\varphi\) as a procedure that generates intervals on the evaluation dataset, i.e., it outputs a sequence of evaluation intervals \( \left( \left\langle \tau^s_1, \tau^a_1, \tau^e_1 \right\rangle, \ldots \right) \).
We also use \(\varphi\) to denote this sequence.
The generated intervals can, e.g., be fixed-length sliding- or tumbling windows.
For any metric \(\sigma\) (e.g., accuracy) and a model \(m_x \in \mathcal{M}_P\), let \(\sigma\left(m_x, \varphi_i\right)\) denote the score of model \(m_x\) on data with timestamps in the \(i\)-th evaluation interval.
We define the evaluation matrix \(m_{\sigma, P} \in \mathbb{R}^{\left| \mathcal{M}_P \right|\times \left| \varphi \right|}\) where, for all \(i \leq \left|\mathcal{M}_P\right|\) and \(j \leq \left| \varphi \right|\), \(m_{\sigma, P}\left[ i,j \right] = \sigma\left(m_i, \varphi_j\right) \).
Each model is evaluated on each window.}

\begin{figure}
        \begin{center}
            \input{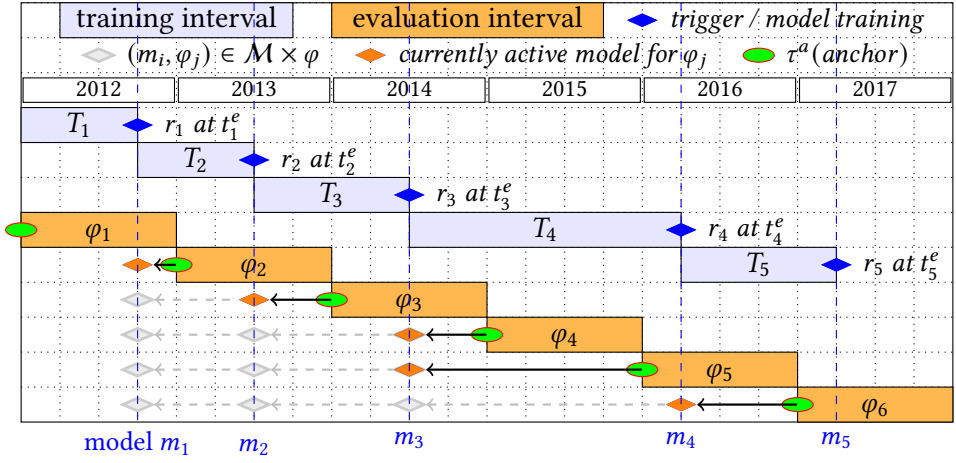}
        \end{center}
    \caption{\revision{Visualization of finding the currently active model.}}
    \label{fig:gantt-active}
    \vspace{-0.5cm}
\end{figure}
 
\revision{\textbf{From matrices to sequence.}
Currently, each pipeline is associated with a 2-dimensional evaluation matrix (models and intervals).
When comparing multiple pipelines, we have to consider another dimension for the pipelines themselves.}
\revision{To reduce the number of dimensions, we propose to define a \emph{composite model} per pipeline.
Formally, the composite model is a partial mapping \( {\mu}_P: \varphi \rightarrow \mathcal{M}_p \).
This allows us to condense the accuracy matrix \(m_{\sigma, P} \in \mathbb{R}^{\left| \mathcal{M}_P \right|\times \left| \varphi \right|}\) into a sequence of evaluation results \(\Lambda_{\sigma, P} = \left( m_{\sigma, P}\left[ \mu_P\left( i \right), i \right] \mid i \leq \left| \varphi \right| \right)\in \mathbb{R}^{\left| \varphi \right|} \).
This sequence represents the temporal performance of a pipeline.
We call it the composite model performance, though the composite model is formally a mapping.
}

\revision{We propose and focus on two variants of composite models.
In the \emph{currently active} composite model, every evaluation window uses the most recent model that has completed training prior to the anchor of the evaluation interval, i.e., \(\mu_P^{active}(\varphi_i) = \arg\max_{m_x \in \mathcal{M}_P} \{t^e_x \mid m_x = \left\langle w_x, \mathcal{D}_x, t^s_x, t^e_x\right\rangle \land t^e_x \leq \tau^a_i \}\).
Intervals whose anchor is before the first model, i.e., when no model training has finished before the evaluation data comes, do not have a currently active model.
It is a modeling decision of the interval generation function whether the anchor point lies on the left boundary of the interval, or, e.g., in the center, to allow for a mix of out-of-distribution and in-distribution data.
}
\revision{\Cref{fig:gantt-active} visualizes this with a Gantt chart of the model and evaluation intervals.
In this example, we set \(\tau^a = \tau^s\).
The training data intervals end with a trigger, indicated by the \textcolor{ganttblue}{blue diamond}.
Conceptually, each \textcolor{light_orange}{evaluation interval} searches (arrows to the left) for the first model that has finished training before its anchor at the beginning of the box.
The model associated with the trigger belonging to the \textcolor{ganttblue}{dashed vertical line} is marked as currently active for the evaluation interval, indicated by the \textcolor{accent_orange}{orange diamond}.
A model can be active for several (\(r_3\)) or no intervals (\(r_5\)).
}

\revision{The \emph{currently trained} composite model is the model following the currently active model.
Let \(i\) be the index such that for the \(j\)-th interval \(\mu_P^{active}(\varphi_j) = m_i\).
We define \(\mu_P^{train}(\varphi_j) = m_{\min(i+1, \left| \mathcal{M}_P \right|)}\).
For the edge case when the currently active model is undefined, we set the most recent model as currently trained.
The currently trained model potentially benefits from training on data distributions similar to those in the evaluation set.
We will see an example of the difference between \(\mu_P^{train}\) and \(\mu_P^{active}\) in~\Cref{sec:evaluation}. 
}
\revision{These definitions emphasize the current performance of a pipeline.
They might not capture other aspects such as retention of previous knowledge.}

\revision{\textbf{Further dimensionality reduction.} For a comparative analysis of pipelines, plotting the temporal accuracy of composite models, i.e., plotting \(\Lambda_{\sigma, P}\), may provide visual insights.
To distill this information into a single metric, the series \(\Lambda_{\sigma, P} \in \mathbb{R}^{\left| \varphi \right|}\) (the composite model accuracy) can be averaged into a pipeline score \(\Sigma_{\sigma, P} \in \mathbb{R}\) to obtain an indication of the general pipeline performance over time.}
\revision{This is how we calculated the mean in~\Cref{fig:motivation-selection} to compare pipeline performance.
Furthermore, this scoring is useful for ranking pipelines in an AutoML setting~\cite{Redyuk2024Pipelines,He2021AutoML}.}

\revision{\textbf{Cost trade-off and pipeline comparison.}
Let \(\mathscr{P}\) be a set of pipelines, each assigned a fixed cost \(P_c\).
Costs can be measured by the number of triggers, the number of samples trained on, or wall clock run time.}
The number of triggers is only fair when all pipelines use the same selection policy on only the new data since the last trigger as in this case each sample from the entire dataset is trained on at most once.
The number of samples is fair across different selection policies but disregards overheads such as the cost of the triggering and selection policies.
Wall clock time covers everything, but requires pipelines to be run on isolated machines.

\revision{
Having assigned a cost, we can build the cost-accuracy feasible set \(\mathbf{F}_\mathscr{P} = \left\{ \left\langle\Sigma_{\sigma, P}, P_c \right\rangle \mid P \in \mathscr{P} \right\}\).
There might be several pareto-optimal pipelines.
For visually comparing pipelines, we can plot this feasible set and get an understanding of how different pipelines perform with respect to the tradeoff between training cost and predictive performance.}

\section{\modyn's Design}\label{sec:design}

\revision{\modyn{} is designed to implement the pipeline model described in~\Cref{sec:formal-modeling}.
Hence, the core unit of execution in \modyn{} is a pipeline.
Users \emph{declaratively specify} the pipeline which allows to decouple the pipeline policy from how it gets executed and lets users focus on model engineering.
}
Still, \modyn{} allows users to add new models and policies as Python modules and offers abstractions to support this (\Cref{sec:implementation}).

\revision{\modyn{} is designed to fill the gap identified in~\Cref{subsec:back-sys}.
To allow users to control which individual data samples to access for training, \modyn's storage component assigns each sample a unique ID and associates metadata with each ID.
Instead of seeing the dataset as a blob of data, \modyn{} offers a \texttt{get\_sample\_by\_id} interface to fetch data according to the selection policy during training.
Next, to support the rich landscape of selection and triggering policies in its declarative interface, \modyn{} introduces a taxonomy of these policies (\Cref{sec:implementation}) and implements abstractions to apply these techniques to common DNN data modalities like text or images.
Furthermore, we design \modyn's rich evaluation infrastructure to support the ideas outlined in~\Cref{sec:formal-modeling}.
}

\begin{figure}
    \centering
    \adjustbox{max width=0.9\linewidth,trim=0cm 0.25cm 0cm 0cm}{%
        \includesvg[width=\linewidth]{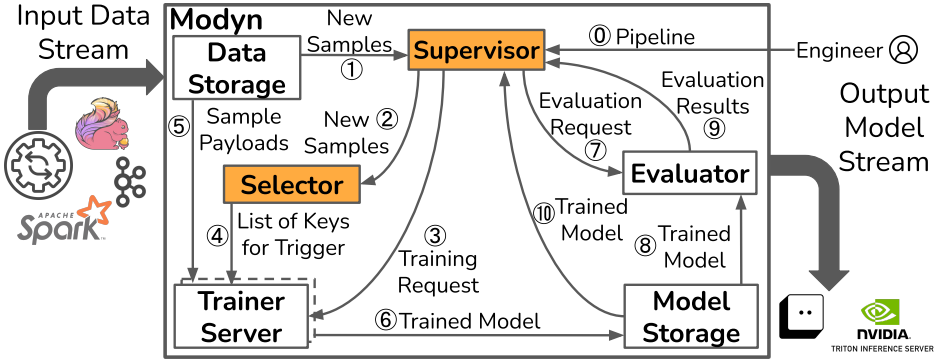}
    }
    \caption{\revision{\modyn's system design.}} 
    \vspace{-0.4cm}
    \label{fig:modyn-arch}
    \Description{TODO!}
\end{figure}

\Cref{fig:modyn-arch} shows \modyn's components and the basic flow of pipeline execution.
\modyn{} is positioned between the preprocessing of the data and the serving of models.
\modyn{} ingests data from a data source, such as stream processing engines (e.g., Flink~\cite{Carbone2015Flink}) or batch processing frameworks (e.g., Spark~\cite{Zaharia2016Spark}). 
\revision{We assume that expensive preprocessing operations, e.g., filtering and downscaling a stream of images, happen offline, i.e., before ingestion into \modyn.
Online ML preprocessing (e.g., image augmentation) happens within \modyn.
}
\revision{While data preprocessing for ML provides challenges in itself~\cite{Xin2021Pipelines}, existing work addresses those challenges~\cite{Schelter2023Argus, Derakhshan2019Retraining,Grafberger2023Pipelines}.
\modyn{} expects a \emph{labeled} input data stream.
Such labels can be either obtained automatically (e.g., track which advertisements a user clicked on) or from human-in-the-loop annotation systems~\cite{Wu2022HITL}.
\modyn{} outputs a stream of trained models that can then be deployed, using tools like TorchServe~\cite{TorchServe}, BentoML~\cite{BentoML}, or Triton Inference Server~\cite{TritonInferenceServer}.}

\textbf{Overview of control flow and data flow.} 
The user submits a pipeline via \modyn's CLI to the supervisor~\Circled{0}, which implements the triggering policy and orchestrates the execution.
\modyn{} stores data samples streaming in from external sources in its storage, which assigns a unique key to each sample.
The data storage component informs the supervisor about new samples by their key~\Circled{1}.
The supervisor checks whether any data point in the incoming batch causes a trigger and forwards potential triggers and the sample keys to the selector~\Circled{2}, which implements the data selection policy.
Upon trigger, the supervisor contacts the trainer server to start a training process~\Circled{3}.
The trainer server requests the trigger training set (keys and weights to train on) from the selector~\Circled{4}.
Then, it loads the actual data from the storage~\Circled{5} and, depending on the configuration, also the previous model from the model storage.
The trainer server then runs a training according to the configuration.
The trained model is then stored in the model storage component~\Circled{6}.
\revision{The supervisor can send an evaluation request to the evaluator~\Circled{7}, which receives the newly trained model from model storage~\Circled{8}, evaluates it and returns the results~\Circled{9}.
The supervisor can also receive the new model for new triggering decisions~\Circled{10}.
}
Finally, the model can be deployed.

\begin{figure}
\begin{minted}
[
frame=lines,
framesep=2mm,
baselinestretch=0.8,
bgcolor=LightGray,
fontsize=\footnotesize,
linenos
]
{yaml}
model:
  id: ResNet18
  config:
    num_classes: 42
data:
  dataset_id: mnist
  transformations: ["transforms.Normalize(...)"]
  bytes_parser_function: |
    def bytes_parser_function(data: memoryview) -> Image:
      return Image.open(io.BytesIO(data)).convert("RGB")
trigger:
  id: DataAmountTrigger
  num_samples: 100
training:
  use_previous_model: True
  batch_size: 1234
  optimizers: ...
  optimization_criterion:
    name: "CrossEntropyLoss"
  selection_strategy:
    name: "CoresetStrategy"
    storage_backend: "database"
    tail_triggers: 0
    presampling_config: ...
    downsampling_config: ...
model_storage:
  full_model_strategy:
    name: "PyTorchFullModel"
  incremental_model_strategy:
    name: "WeightsDifference"
evaluation: ...
\end{minted}
\vspace{-0.75cm}
\caption{Excerpt from an example \modyn{} pipeline.}
\Description{TODO!}
\label{listing:modyn-pipeline}
\end{figure}

\textbf{Example pipeline.}
\revision{\Cref{listing:modyn-pipeline} shows a declaratively-specified \modyn{} pipeline.
At minimum, a description comprises} \textit{(1)} the model specification, \textit{(2)} the training dataset and a corresponding bytes parser function that defines how to convert raw sample bytes to model input, \textit{(3)} the triggering policy, \textit{(4)} the data selection policy, \textit{(5)} training hyperparameters such as the the learning rate and batch size, \textit{(6)} training configuration such as data processing workers and number of GPUs, and \textit{(7)} the model storage policy, i.e., a definition how the models are compressed and stored.
\revision{A training may involve fine-tuning a model or training a model from scratch with randomly initialized weights; this is a configuration parameter in the triggering policy.
The very first training can run on a randomly initialized or externally provided model.}

\section{Implementation}\label{sec:implementation}

We describe the supervisor and triggering policies (\Cref{subsec:impl-supervisor}), the selector and data selection policies (\Cref{subsec:impl-selector}), data retrieval (\Cref{subsec:impl-fastretrieval}), and the remaining components (\Cref{subsec:impl-other}).
We build \modyn{} with the goal of providing an easy-to-use, extensible, and efficient execution platform for data-centric ML pipelines. 
We aim to build an ecosystem around \modyn{} to facilitate policy exploration in practical use cases of ML training in growing data environments.

To balance performance and ease-of-use, \modyn{} components are either written in C++ (e.g., storage service), purely in Python (e.g., trainer service), or Python with C++ extensions (e.g., selector service).
While code on the hot path of data fetching is written in C++ to avoid stalls, the pluggable algorithm modules are written in Python.
Having a clean Python interface allows ML researchers to implement policies in a familiar language without worrying about systems aspects.
For compatibility, we use existing tooling like PyTorch where possible.

\modyn{} uses gRPC and FTP for data and control flow, and supports Docker Compose for deployment.
The codebase, totaling ca. 20\,000 lines of Python and 2\,500 lines of C++ (excluding tests), is publicly accessible\footnote{Available at \mbox{\url{https://github.com/eth-easl/modyn}}.}, and undergoes rigorous unit and integration testing, as well as linting, establishing it as more than a research prototype.

To overcome the limitations imposed by the Global Interpreter Lock (GIL) in Python, our implementation employs a hybrid processing and threading approach. 
It utilizes a gRPC ThreadPool and \texttt{multiprocessing.Processes}, leveraging the \texttt{SO\_REUSEPORT} socket option. 
This combination enables the system to handle multiple gRPC requests concurrently, achieving true parallelism despite the GIL constraints.

\subsection{Supervisor}\label{subsec:impl-supervisor}
The supervisor orchestrates the execution of pipelines.
\revision{Pipelines are submitted via \modyn's CLI.
The CLI is the interface between supervisor and user.
\modyn{} uses Pydantic models~\cite{Pydantic} to guide users in specifying their pipelines.}
For each submitted pipeline, the supervisor spawns a \texttt{PipelineExecutor}, which implements a state machine following the control flow outlined in~\Cref{sec:design}.
The client frequently polls the supervisor for the current status and displays the current pipeline stage and training progress.

\textbf{Triggering policies.}
\revision{During execution, the supervisor decides to trigger using a triggering policy.}
\modyn{} currently supports amount-, time-, \revision{performance-,} and drift-based triggering policies.
Amount triggers fire every \(n\) data points, while time triggers fire after a time interval has passed.
\revision{Performance triggers trigger when the accuracy degrades.
They require labels which might arrive late in practice~\cite{Shankar2022Observability}.
Drift triggers, however, work unsupervised and detect \emph{covariate shift}, i.e., they compare the distribution of the incoming data to some reference data.
We leverage the evidently~\cite{evidently} and alibi-detect~\cite{alibi-detect} libraries for calculating similarity metrics and hypothesis testing.
}

\textbf{Data drift variants.}
\revision{For unstructured data such as images or text, we first need to transform it into a latent embedding space, and optionally project it to lower dimensionality, e.g., using principal component analysis (PCA).}
\modyn{} uses the most recent model of the pipeline to generate embeddings.
\modyn{} builds up a sliding window of current data and reference data (current data window at the last trigger).
In a defined interval, a similarity metric between the two windows is obtained.
Based on the similarity metric, we need to make a binary decision about whether there is drift between the reference and current data, i.e., whether we should fire a trigger.
\modyn{} supports threshold-based decisions, i.e., we trigger when the metric is higher than a configurable threshold.
\revision{As this threshold needs to be tuned for each dataset and metric, \modyn{} also supports dynamic decision making (\texttt{AutoDrift}).
In this setting, \modyn{} keeps track of a window of previously observed drift scores and triggers when a new drift score is in a configurable percentile of these scores as a simple outlier detection.
}

\revision{\textbf{Similarity metrics.}
While tabular data, as used in previous work~\cite{Werner2024DriftPerformance,Redyuk2021DQV,Schelter2020Validate}, can be used directly, for images and text, \modyn{} generates embeddings as dense latent representations, and calculates drift metrics on those embeddings.
The embedding dimensions become features as in the tabular data domain. 
We find that some distance metrics, such as the Kolmogorov–Smirnov or Hellinger distance~\cite{Ditzler2011Hellinger},  are commonly used for univariate distributions.
In univariate drift detection, we derive one distance metric \emph{per feature}, e.g., for 512-dimensional embeddings, we obtain 512 distance values that need to be reduced into a scalar.
Multivariate extensions or natively multivariate metrics provide a scalar distance value, even for multivariate distributions.
We focus on the multivariate maximum mean discrepancy (MMD) metric~\cite{Gretton2012TST} since we did not find readily available multivariate implementations of other metrics.
Additionally, it has not been explored how to best reduce multiple univariate metrics into a scalar value, how to decide whether the data has drifted, and MMD performed best in initial experiments.}

\textbf{Open questions.}
Using drift detection on unstructured data such as images is an active area of research.
First, the impact of the embedding space, i.e., which model is used to generate embeddings, has not been explored.
Second, it has not been studied what is a sensible interval to run detection, what metric to choose in which scenarios, and how big the windows should be.
Last, it is not clear what is the best way to make the binary triggering decision, and it likely depends on the metric, dataset, embeddings, etc.
Note that our goal is to demonstrate how \modyn{} enables the use and exploration of different triggering policies rather than advocating for a particular policy.
We are actively exploring these questions and discuss the first results in~\Cref{subsec:eval-drift}.

\textbf{Execution modes.} 
\modyn{} advocates the principle of \emph{what you evaluate is what you deploy}.
Managing separate codebases for research and production is error-prone. 
Hence, any pipeline can be executed in either \emph{experiment mode} or \emph{production mode}.
In production mode, the data storage informs the supervisor when new data points arrive.
In experiment mode, the data storage simulates new data points streaming in by announcing existing data points as \enquote{new} to the supervisor.
The experiment mode can be used to (re)play traces and compare how policies perform given the same environment.
The insights gained from these experiments can then be used to find a configuration for production.

\subsection{Selector}\label{subsec:impl-selector}
\revision{The selector implements data selection policies, which generate the trigger training set \(\mathcal{D}_r\) upon the \(r\)-th trigger per pipeline.}

\subsubsection{Selection policies}\label{subsubsec:impl-selector-policies}
A selection policy defines what data to train a model on upon trigger.
Every selection policy has a window upon the past data, i.e., a pool of data we could train on.
This window can be infinite (\enquote{retrain} on all past data), just include the data since the last trigger (\enquote{finetune} on new data), or include all data up to \(n\) previous triggers.
In order to either reduce the amount of data that we train on or increase information retention, we can then apply selection algorithms on this window of data.
\revision{In the following, we discuss a taxonomy of selection policies.}

\textbf{Presampling and downsampling.}
\revision{We identify two types of selection algorithms.
Presampling algorithms do not require any information from the model forward pass and are implemented at the selector.}
Examples include ingesting older samples to increase information retention, sampling in a class-balanced fashion, or use-case specific sampling (e.g., increasing the weight of pictures at night for autonomous driving pipelines). %

Downsamplers are general-purpose techniques which leverage information from the model forward pass to pick the best samples to use for the backward pass~\cite{Mirzasoleiman2020CRAIG,Pooladzandi2022AdaCore,a,c}.
Downsampling happens at the trainer server.
For example, the \textsc{DLIS} policy~\cite{a} samples data points based on the gradient norm obtained during the forward pass.
Any downsampler can be combined with an offline or online presampling policy.

\begin{figure}
    \centering
    \adjustbox{max width=0.7\linewidth,trim=0cm 0.55cm 0.85cm 0cm}{%
        \includesvg[width=\columnwidth]{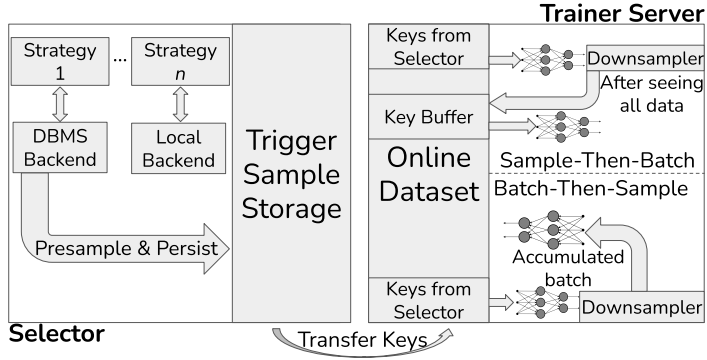}
    }
    \caption{Data selection flow in \modyn.} 
    \label{fig:selector-ts}
    \vspace{-0.5cm}
    \Description{TODO!}
\end{figure}

\textbf{Offline/online presampling.} 
Any presampling policy is offline or online.
Offline policies maintain state by storing all samples during a trigger and running the actual selection on trigger.
For example, a strategy sampling class-balanced from the data window requires storing all data first and only samples on trigger after determining the available classes.
Online policies perform the sampling directly as data is received.
Examples for online policies include continual learning algorithms such as \textsc{GDumb}~\cite{Prabhu2020GDumb}, \textsc{CLiB}~\cite{Koh2022CLIB}, and \textsc{GSS}~\cite{Aljundi2019GSS}.

\textbf{Supported policies.}
Currently, for presampling, \modyn{} supports class-balanced sampling (similar to \textsc{GDumb}~\cite{Prabhu2020GDumb}), sampling uniformly at random, and trigger-balanced sampling.
For downsampling, \modyn{} supports RS2~\cite{b}, loss sampling~\cite{a}, \textsc{DLIS}~\cite{a}, uncertainty downsampling~\cite{c}, \textsc{CRAIG}~\cite{Mirzasoleiman2020CRAIG}, and \textsc{GradMatch}~\cite{Killamsetty2021GradMatch}.
It also implements a warmup period of not using sampling for the first triggers to improve upon the initial model more quickly.

\subsubsection{Implementation of policies.}
Presampling and downsampling policies are implemented as Python classes, each category sharing its own common interface.
\modyn{} provides infrastructure, e.g., for storing state, to help engineers and researchers port algorithms.
The overall flow of data selection is shown in~\Cref{fig:selector-ts}, which we detail in the following paragraphs.
When informed about new samples, the selection policy updates its state using a \emph{metadata backend} module provided by the selector.
This state is used to calculate the set \(\mathcal{D}_r\) on trigger \(r\).
This set is then stored on disk using an extension called \texttt{TriggerSampleStorage} (\Cref{subsec:impl-fastretrieval}).

\textbf{Backends for presampling.}
For implementing presampling strategies, \modyn{} provides two backends that share an interface to store the state of the sampling strategy.
The first backend is the Postgres backend, which persists the samples to a Postgres table~\cite{Stonebraker1986Postgres}.
The advantage of this backend is the flexibility for implementing selection policies, since many policies can be expressed using SQL statements.
We use SQLAlchemy~\cite{sqlalchemy} to allow for easy querying of data.
\modyn{} provides query boilerplates using inheritance hierarchies, e.g., in order to implement a random sampling balanced across some parameter such as trigger or label, the developer inherits from the \texttt{AbstractBalancedStrategy} and specifies the column to balance on.
The disadvantage of the Postgres backend is the slow insertion speed.
Every sample has to be written into the database.
We optimize the ingestion with Postgres' table partitioning mechanism.
We partition the state table first by pipeline, then by trigger, and then round-robin with a modulus of 16.
This avoids the degrading of insertion performance with growing number of triggers, since every trigger defines a new physical table.
In order to further optimize the insertion speed, we use SQL bulk insertion and run several insertion threads for new batches of incoming keys.

For datasets with many samples, e.g., recommendation system datasets, using Postgres can be very expensive.
We observe maximum insertion speeds of around 100\,000 insertions/second.
For simple strategies not requiring complex SQL queries (e.g., train on all the data since the last trigger), or if performance is key, \modyn{} offers a local backend.
This is a C++ extension that writes data multithreadedly to a local disk, such as a high performance NVMe drive.
These binary files are written and read avoiding unnecessary memory copies.
Strategies such as training on all data, uniform presampling, or mixing old and new data can be implemented easily on this backend, trading off ease of implementation for speed.
Each workload has different requirements and \modyn{} provides building blocks for these use cases.

\textbf{Implementing downsamplers.}
Downsampling policies cannot be executed at the selector and need support from the trainer server.
\modyn's training loop has a component which executes the downsampling policy specified in the pipeline.
As shown in~\Cref{fig:selector-ts}, the presampled trigger training set is transferred to the trainer server, where it is then downsampled.

Analogous to offline versus online presampling, downsamplers can be run in either \emph{sample-then-batch} (StB) or \emph{batch-then-sample} (BtS) mode.
Some downsamplers like RHO-LOSS~\cite{Mindermann2022RHOLOSS} explicitly are proposed with BtS or StB mode, and others like \textsc{DLIS}~\cite{a} can be used in both modes.
In BtS, the training loop runs inference on a batch and then selects a subset of that batch.
This is repeated until we accumulate a new batch of the original batch size, on which we then perform a backward pass.
In StB mode, the training loop starts with a sampling phase in which it continuously informs the downsampler about the forward pass, allowing the downsampler to build up state for all samples.
\revision{Once this state is complete, it generates the downsampled dataset, and we run training on these keys.}
This sampling phase can be performed every training epoch or less often.
Both StB and BtS mode are abstracted such that engineers just have to implement one version of the downsampling policy.
While there can be multiple epochs of training per trigger, \modyn{} applies the budget constraint per epoch.
For example, when we have 1\,000 samples, a 10\,\% budget, and 10 epochs per trigger, we train for 10 epochs of 100 samples each, instead of a single epoch of 1\,000 samples.
Maintaining the epoch boundary allows, e.g., consistent setups for learning rate scheduling.

\subsection{Fast Data Retrieval}\label{subsec:impl-fastretrieval}
\modyn{} supports different data selection policies, which means that the trigger training set is an arbitrary collection of previously stored samples.
Regardless of the selection policy, the result is a list of training items, i.e., IDs of samples, to train on.
This is a shift in architecture from traditional ML deployments, where the training data is typically a big chunk of data that can be read sequentially.
Instead, \modyn{} supports \emph{sample-level data selection}, i.e., retrieving samples based on their identifier.
For big datasets with potentially billions of small samples like in recommendation systems, this can lead to data stalls during training.
In this subsection, we describe how we engineer \modyn{} to avoid data stalls while supporting sample-level data selection.
We first describe the storage component (\Cref{subsec:impl-storage}). 
Then we describe the \texttt{OnlineDataset} abstraction that loads keys from the selector, payloads from storage, parses the bytes, and returns tensors into the training loop (\Cref{subsubsec:impl-onlinedataset}).
We furthermore explain how the selector quickly returns the list of keys (\Cref{subsubsec:impl-tss}), and how, given a list of keys, the storage quickly returns the requested data (\Cref{subsubsec:impl-storage-dataretr}).

\subsubsection{Data Storage}\label{subsec:impl-storage}
The storage is entirely written in C++ as we found the data wrangling to be particularly expensive in Python.
A Postgres database is used to keep track of all available samples.
Each ingested file can contain one or more samples, e.g., a JPEG file contains exactly one sample, while a CSV file contains potentially hundreds of thousands of samples.
When the component encounters a new file, it extracts all the samples in that file and inserts the file, the sample IDs, and the labels into the database.
The storage makes use of \texttt{FileSystemWrappers} which abstract I/O operations such as reading byte streams from files. %
Currently, \modyn{} implements a file system wrapper for the local file system, but this can be easily extended to support cloud file systems like S3.
The storage then uses \texttt{FileWrappers} which abstract how to extract individual samples and labels from files.
Examples include the \texttt{CSVFileWrapper} for variable-length CSV data, the \texttt{BinaryFileWrapper} for fixed-size columnar data, often used in recommendation systems training, and the \texttt{SingleSampleFileWrapper} for files containing exactly one sample, such as images.

The C++ implementation uses SOCI~\cite{SOCIRepo} to operate on the Postgres database.
To optimize the ingestion and query performance, we partition the tables.
Since for datasets with billions of samples, even SQL bulk insertion is too slow, we use the Postgres internal \texttt{COPY} command and stream the data over the raw connection.

\subsubsection{The \texttt{OnlineDataset}}\label{subsubsec:impl-onlinedataset}
The \texttt{OnlineDataset} abstracts away the interaction with the different gRPC components from the training loop.
The training loop (\Cref{subsec:impl-other}) uses a standard PyTorch DataLoader to fetch batches to train on.
It is not aware of the ongoing network communication.
This new abstraction is necessary due to \modyn's sample-level data selection.
We cannot just load a big chunk of data and train on it.
Instead, we have to load the data according to the list of keys in the trigger training set.
The PyTorch DataLoader uses multiple workers. 
We split the trigger training set across these workers.
The trigger training set consists of fixed-size partitions (\Cref{subsubsec:impl-tss}).
Each worker gets an equal share of each partition.
The data loader fetches batches from the workers in a round-robin fashion.

\begin{figure}
    \centering
    \adjustbox{max width=0.8\linewidth,trim=0cm 0.4cm 0.5cm 0.2cm}{%
        \includesvg[width=\columnwidth]{img/dataloading.svg}
    }
    \caption{The architecture of the \texttt{OnlineDataset}.}
    \label{fig:dataloading}
    \vspace{-0.65cm}
    \Description{TODO!}
\end{figure}

In order to avoid data stalls when the data loader requests data, each worker (or dataset instance) implements a prefetching mechanism.
This architecture is depicted in~\Cref{fig:dataloading}.
Each worker has a partition buffer of a configurable size.
Upon creation, a worker spawns a configurable number of prefetching threads that issue gRPC requests.
The size of the buffer defines how many partitions we prefetch overall, while the number of threads defines how many partitions we prefetch in parallel.
To fetch a partition, we first obtain a list of keys from the selector, and then ask the storage for the payloads corresponding to these keys.
The storage uses gRPC streaming to transfer the payloads to the workers.

As soon as data is available in the buffer, the main thread of the worker fetches the payload, applies transformations, and yields it to the data loader.
This is important, since waiting for a partition to finish transferring would make the batch latency depend on partition size.
The only exception is when if shuffling is enabled, i.e., we need to shuffle the samples in each partition and the order of partitions, as we need to alter the sample order.
The first transformation always is a user-defined \emph{bytes parser function} defining how to transform the bytes of the payload to a tensor, e.g., by decoding the bytes of a JPEG image or decoding a UTF-8 string.
Afterwards, other transformations are applied as defined by the pipeline, such as image augmentations or tokenization. %

\subsubsection{Data partitioning}\label{subsubsec:impl-tss}

We need to retrieve the trigger training set, i.e., the keys to train on, as fast as possible.
Instead of relying on a database, we persist the fixed trigger training set after presampling to disk using the \texttt{TriggerSampleStorage} (TSS).
The TSS is a fast C++ extension that persists the list of keys and weights (c.f.~\Cref{sec:formal-modeling}) output by the presampling strategy to disk.
The TSS uses the same binary file format as the local backend.

\textbf{Writing to disk.}
The selection strategy does not pass all keys and weights at once to the TSS.
Instead, it passes the keys as multiple partitions.
Each partition is a fixed-size set of keys.
For example, if the trigger training set consists of 1\,000 keys and the partition size is 100, the strategy will pass 10 partitions to the TSS.
This avoids high memory utilization by limiting the amount of keys loaded at once.
Furthermore, the partitions provide a fixed-size unit of data transfer for the trainer server.
The backends provide support for partitioning, i.e., limiting the memory usage.
For the Postgres backend, we use Postgres' server-side cursors.
For the local backend, we read the corresponding data via offsets.
When the TSS writes the final partition to disk, \(n\) threads (within the C++ extension) write the keys and weights of the partition to disk in parallel.

\textbf{Retrieving keys.}
When retrieving partition data for a worker, we iterate over all files for this partition.
The requesting worker ID and the number of total samples correspond to a list of samples for this worker.
However, as the number of dataloader workers does not necessarily match the number of threads we used to persist the training set to disk, we have to potentially parse subparts of files and correctly and efficiently assemble each worker's share of a partition.
This is hidden in the C++ extension and only the final list of keys is returned.

\subsubsection{Storage data retrieval}\label{subsubsec:impl-storage-dataretr}

What makes the storage challenging is that it can receive requests with arbitrary sets of sample keys.
When samples are requested, they are distributed across a set of files, and each may be residing at arbitrary locations within those files.
The storage needs to efficiently build a buffer of data that makes it look as if the data came from one continuous file that contained all requested samples.

When a worker sends a list of keys to the storage for retrieval, the storage partitions this list into \(n \geq 1\) parts to parallelize the retrieval from disk.
Then, each thread obtains labels and a source file for each sample from Postgres, grouped by file.
For each file, it instantiates a \texttt{FileWrapper} and extracts all samples in that file into a send buffer.
When that buffer is full, or once all files have been iterated through, the thread emits the buffer to the worker.
Besides parallelization, major speed gains for each thread stem from optimized \texttt{FileWrapper} implementations.
For example, the \texttt{BinaryFileWrapper} has an optimized bytes-to-int parsing function based on the endianness the file was written with, and operates on \texttt{std::ifstream}s to not load the entire file into memory.

\subsection{Other Components}\label{subsec:impl-other}

\subsubsection{Trainer Server}\label{subsec:impl-ts}
The trainer server spins up trainers when requested, which execute a general-purpose training loop.
\modyn{} currently implements a PyTorch-based trainer, but its design is agnostic to the ML framework. %
The trainer supports a variety of features like mixed-precision training or learning rate schedulers with correct support for data selection~\cite{b}.
\modyn{} comes with some models (e.g., ResNets~\cite{He2016ResNet}, DLRM~\cite{Naumov2019DLRM,DLRMNvidia}, and transformers~\cite{Wolf2020Transformers}) and other models can be added easily.
\revision{The trainer also performs online featurization, such as image augmentation.}

\subsubsection{Model Storage}
This component is responsible for model storage and retrieval.
\revision{It supports full model and incremental compression policies.}
The full model policy defines how to compress the entire model such that it can be restored from just the file itself, analogous to an I-frame in video encoding.
Furthermore, the model storage can employ an incremental policy, which activates a configurable number of times between full model steps.
In this mode, \modyn{} stores just the delta from the base model based on a specified difference operator.
This is similar to a P-frame in video encoding.
For full model policies, the model storage currently supports both the native PyTorch format and a custom, stripped binary storage format, with or without zip compression.
For incremental policies, it currently supports an xor and subtraction based difference operator.
Model compression over time is an active area of research~\cite{Strassenburg2023MMM,Hao2024Models}.

\subsubsection{Evaluator.}
\revision{Each model trained during a pipeline can be evaluated on several evaluation intervals for multiple evaluation metrics.
\modyn's evaluator implements various interval generation functions \(\varphi\), e.g., tumbling- or sliding windows.
It also supports both decomposable (e.g., accuracy) and holistic metrics (e.g., ROC-AUC).
}

\section{Benchmark Suite}\label{subsec:design-benchmarks}

A major hurdle for research on growing datasets is the scarcity of publicly accessible datasets that encapsulate temporal dynamics and distribution shifts. 
\modyn{} incorporates a benchmark suite that curates datasets, pipeline configurations, and models to run pipelines with.
It comes with the necessary tooling for making them available on the user's machine as some datasets involve post-processing and metadata scraping.
The suite includes:
\begin{enumerate}[leftmargin=*,nosep]
\item The \textsc{Wild-Time} benchmarking suite~\cite{Yao2022WildTime}: A compilation of five datasets, ranging from small to medium in size, each exhibiting distribution shifts. %
\item Kaggle \textsc{arXiv} and \textsc{HuffPost} datasets: The \textsc{arXiv} and \textsc{HuffPost} datasets from Wild-Time only have coarse-grained timestamps on a year resolution and have been filtered by unclear criteria. 
\modyn{} provides tooling to generate full, high resolution versions using the source data from Kaggle~\cite{KaggleArXivDataset,misra2022news}.
\item The \textsc{Criteo} 1\,TB dataset~\cite{Criteo2013Dataset}: The \textsc{Criteo} click stream dataset for recommendation systems training provides user data over 24\,days, with roughly 180\,million samples per day.
\item The \textsc{CGLM} dataset(s)~\cite{Prabhu2023CGLM}: The paper on \textsc{CGLM} classifies images from landmarks on Wikipedia and uses the upload timestamps. Since the original data is not accessible, \modyn{} provides an open-source reproducible script and pre-scraped metadata to generate different versions of the \textsc{CGLM} dataset, e.g., by using the clean or non-clean and hierchical or non-hierarchical version of the original (non-continual) \textsc{CGLM} dataset~\cite{Weyand2020GLM,Ramzi2023GLMHierarchical}.
\item The \textsc{CLOC} dataset~\cite{Cai2021CLOC}: \textsc{CLOC} is a big continual learning dataset on images with distribution shift. \modyn{} supports the version processed by Hammoud et al.~\cite{Hammoud2023rapid}.
\end{enumerate}

While such data often is business-critical, to facilitate future research, we call for more datasets with distribution shift to be released.
Releasing such datasets can help research to solve meaningful problems for practice. %
\modyn{} comes with tooling for analyzing pipelines.
It provides an interactive dashboard based on Dash/Plotly that allows users to (a) analyze single pipelines, i.e., dive into the model and system metrics, and (b) compare pipelines to understand which policies perform best.
Most plots in this paper have first been explored using this dashboard.

\section{Evaluation}\label{sec:evaluation}
We evaluate \modyn{} to answer the following three questions:

\begin{enumerate}[leftmargin=*,nosep]%
    \item How do data selection policies influence accuracy?   
    \item How do different  triggering policies compare? In particular, can drift triggers be used to reduce pipeline cost while keeping accuracy?
    \item What is the impact of \modyn's parallelism, partitioning, and prefetching optimizations and how should the corresponding parameters be set to maximize throughput? How does the per-sample data ingestion throughput compare to reading data sequentially from local storage?
\end{enumerate}

\noindent For all experiments, we use a server with two 16 Core AMD EPYC 7313 CPUs, 256\,GB  DRAM, a 4\,TiB Samsung MZQL23T8HCLS NVMe, and a NVIDIA RTX 3090 GPU.
We use gRPC 1.64.1, Postgres 15.2, PyTorch 2.2.1, NVIDIA GPU driver 545.23.06 with CUDA 12.3, on Ubuntu Server 22.04 with kernel 5.15.
\modyn{} is compiled with GCC 12 and \texttt{-O3 -march=native}.

\subsection{Data Selection}\label{subsec:eval-selection}
In this subsection, we explore the impact of data selection policies on pipeline accuracy.
Each data selection policy needs to define a window of data, a presampling, and a downsampling policy.
We pick the \enquote{finetuning} setting, i.e., we finetune the model from the previous trigger and set our window to contain the data since the last trigger.
We mostly focus on downsamplers (in BtS mode) because they do not require domain-specific knowledge and are built for increasing accuracy on the current distribution.

Due to space constraints, we consider the \textsc{yearbook} dataset~\cite{Yao2022WildTime} and the \textsc{CGLM-landmark} dataset.
We run all pipelines on three seeds and average the results.
We shuffle the training data and use the currently trained composite model.
We test presampling uniform at random (\textsc{uniform}, which samples a subset once and then trains on that for several epochs), class-balanced presampling, RS2 with and without replacement~\cite{b}, \textsc{loss} downsampling~\cite{a}, \textsc{DLIS} downsampling~\cite{a}, and the \textsc{margin}, \textsc{least conf.}, and \textsc{entropy} variants of uncertainty downsampling~\cite{c}.
All policies are implemented in less than 130 lines of code.

\subsubsection{\textsc{yearbook} dataset}\label{subsubsec:eval-selection-yearbook}

\begin{figure}
    \centering
    \begin{adjustbox}{trim=0cm 0.4cm 0cm 0cm}
        \includesvg[width=0.65\linewidth]{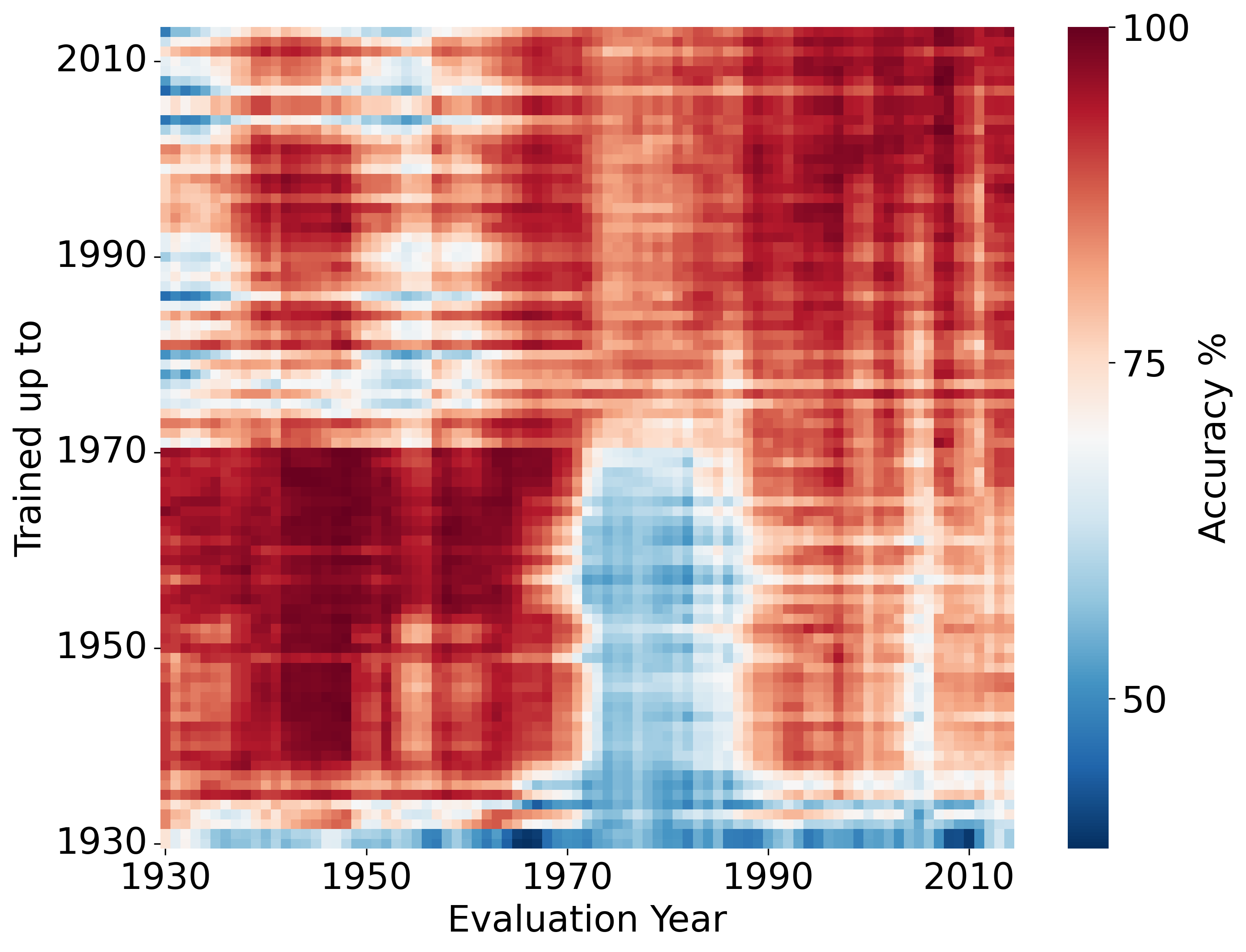}
    \end{adjustbox}
    \caption{Accuracy matrix for \textsc{yearbook} full data training.} 
    \label{fig:yearbook-matrix}
    \Description{TODO!}
\end{figure}

The \textsc{yearbook} dataset classifies school yearbook pictures from 1930 to 2013.
We follow Yao et al.~\cite{Yao2022WildTime} and use their \enquote{yearbooknet} CNN and the training hyperparameters with a batch size of 64, SGD with a learning rate of 0.001, and momentum 0.9.
We also use their evaluation split.
We trigger yearly, i.e., with the highest resolution possible for this dataset, train for 5 epochs per trigger, and use two warmup triggers where we do not apply data selection.
\revision{Due to the small dataset size (33\,431 training  samples), we use a three year sliding window as an interval generation function (\Cref{sec:formal-modeling}) to smoothen the accuracy curve and only run 50\,\% subset selection.}

\textbf{Full data training.}
\revision{In~\Cref{fig:yearbook-matrix}, we show the accuracy matrix \(m_{\sigma, P}\) of full data training on \textsc{yearbook} that we seamlessly obtain using \modyn's evaluation support.}
In the 1970s, we observe a drop in accuracy for models trained on data before this period, in-line with numbers from Yao et al.~\cite{Yao2022WildTime} (Figure 4a), indicating distribution shift.
We hypothesize that, e.g., changing hairstyles over the decades could cause the shift.
As expected, the highest accuracies lie on the diagonal of the matrix, and we can that see the first models underfit.
The low accuracies in the upper left area show how models trained on newer data forget the past distribution.

\begin{figure}[b]
    \centering
    \begin{adjustbox}{trim=0cm 0.5cm 0cm 0cm}
    \includesvg[width=0.75\linewidth]{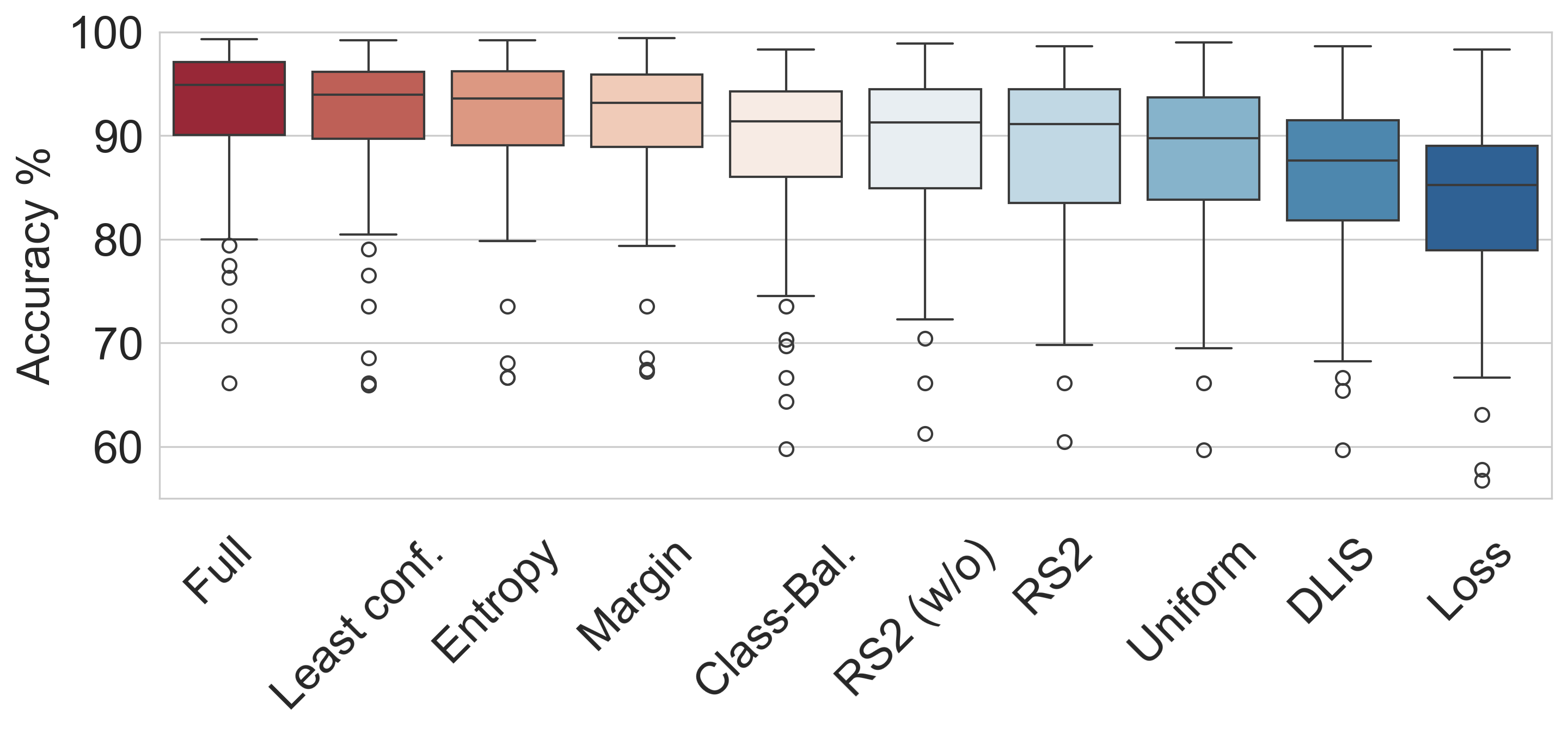}
    \end{adjustbox}
    \caption{Currently trained composite model accuracies for full data training and 50\,\% data selection on \textsc{yearbook}.} 
    \label{fig:benefit-selection}
    \vspace{-0.5cm}
    \Description{TODO!}
\end{figure}

\textbf{50\,\% subset training.}
\Cref{fig:benefit-selection} shows composite model accuracies per selection strategy in a boxplot.
Generally, the uncertainty based downsamplers~\cite{c} perform best.
Full data training has an average accuracy \revision{(pipeline score \(\Sigma_{\sigma, P}\))} of 92.3\,\%, and with 50\,\% selection, \textsc{entropy} reaches 91.4\,\%, and \textsc{least conf.} and \textsc{margin} reach 91.2\,\%. 
\textsc{RS2}~\cite{b} reaches 88.8\,\% (w/o replacement)/88.4\,\% (w. replacement).
\textsc{Loss} and \textsc{DLIS} perform worse than uniform and class-bal. sampling \revision{on this dataset.}

\begin{figure}
    \centering
    \begin{adjustbox}{trim=0cm 0.5cm 0cm 0cm}
        \includesvg[width=0.75\columnwidth]{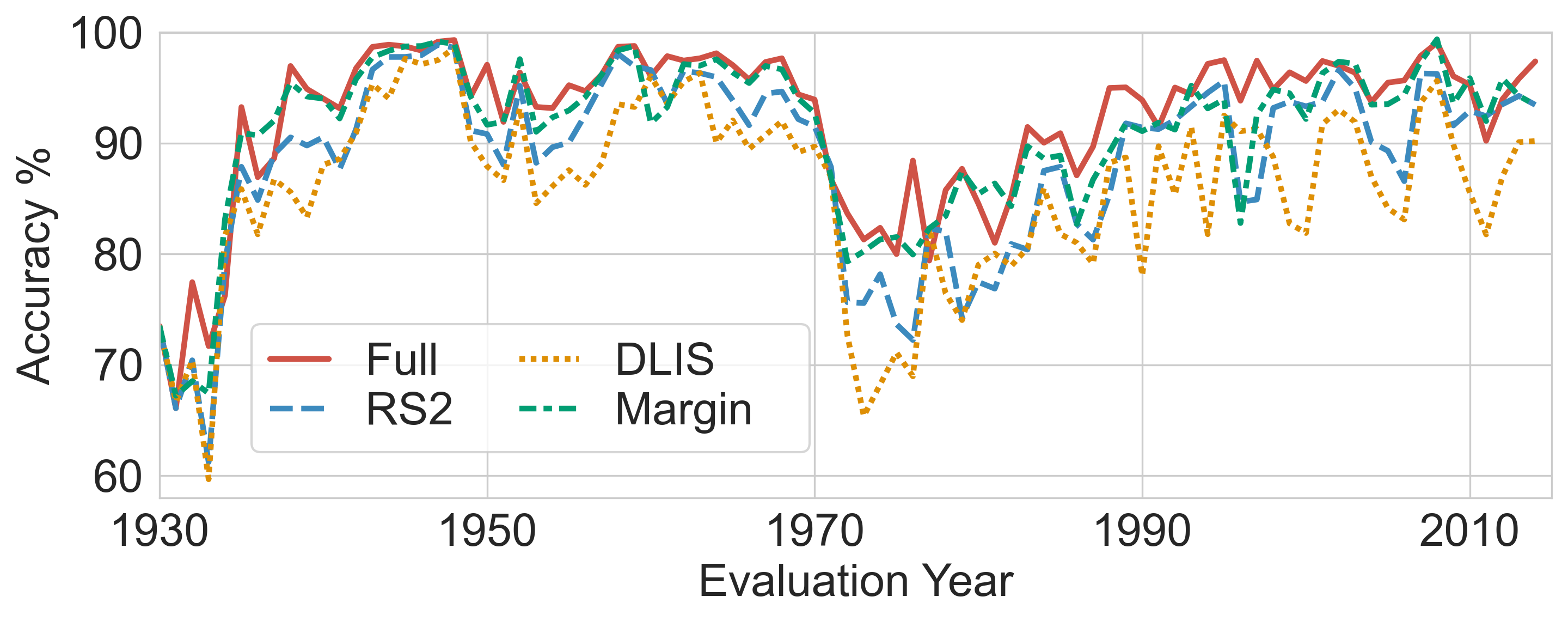}
    \end{adjustbox}
    \caption{Composite model accuracy over time for \textsc{DLIS}, \textsc{margin}, and \textsc{RS2} (w/o) on \textsc{yearbook}.} 
    \label{fig:yearbook-overtime}
    \Description{TODO!}
\end{figure}

We investigate why the average accuracy is higher in~\Cref{fig:yearbook-overtime}.
\textsc{DLIS}'s performance degrades during the drift period, while \textsc{margin} is able to handle the drift better, similar to full data training.
It is able to identify which data points are the most relevant during the shift.
Overall, we find that with uncertainty-based downsamplers we almost reach full-data model accuracy with a 50\,\% training budget.

\subsubsection{\textsc{CGLM-landmark} dataset}\label{subsubsec:eval-cglm}

This dataset classifies pictures from Wikipedia into 6\,404 landmark classes.
We follow Prabhu et al.~\cite{Prabhu2023CGLM} without filtering out uncleaned data, as downsampling might help to recognize unclean data.
Despite applying weaker filter criteria, we obtain 361\,671 samples before splitting the evaluation set, while Prabhu et al.~\cite{Prabhu2023CGLM} claim to obtain 430\,K/580\,K images (they mention both numbers).
Since their data preprocessing is not public, we cannot investigate the differences.
Following Prabhu et al.~\cite{Prabhu2023CGLM}, we train a ResNet50~\cite{He2016ResNet} with pretrained weights from ImageNet, and use SGD with a learning rate of 0.005 and momentum of 0.9.
We use a batch size of 128 and train for 5 epochs per trigger.
We trigger every year.
Since the first years contain very little data, we use 5 warmup triggers.
We evaluate using one year tumbling windows, and report top-5 accuracy since this is a hard classification task with 6\,404 classes.
We filter out the years 2005, 2006, and 2020 due to the low number of samples in the evaluation set.

\begin{figure}[b]
    \centering
    \begin{adjustbox}{trim=0cm 0.3cm 0cm 0cm}
        \includesvg[width=0.75\columnwidth]{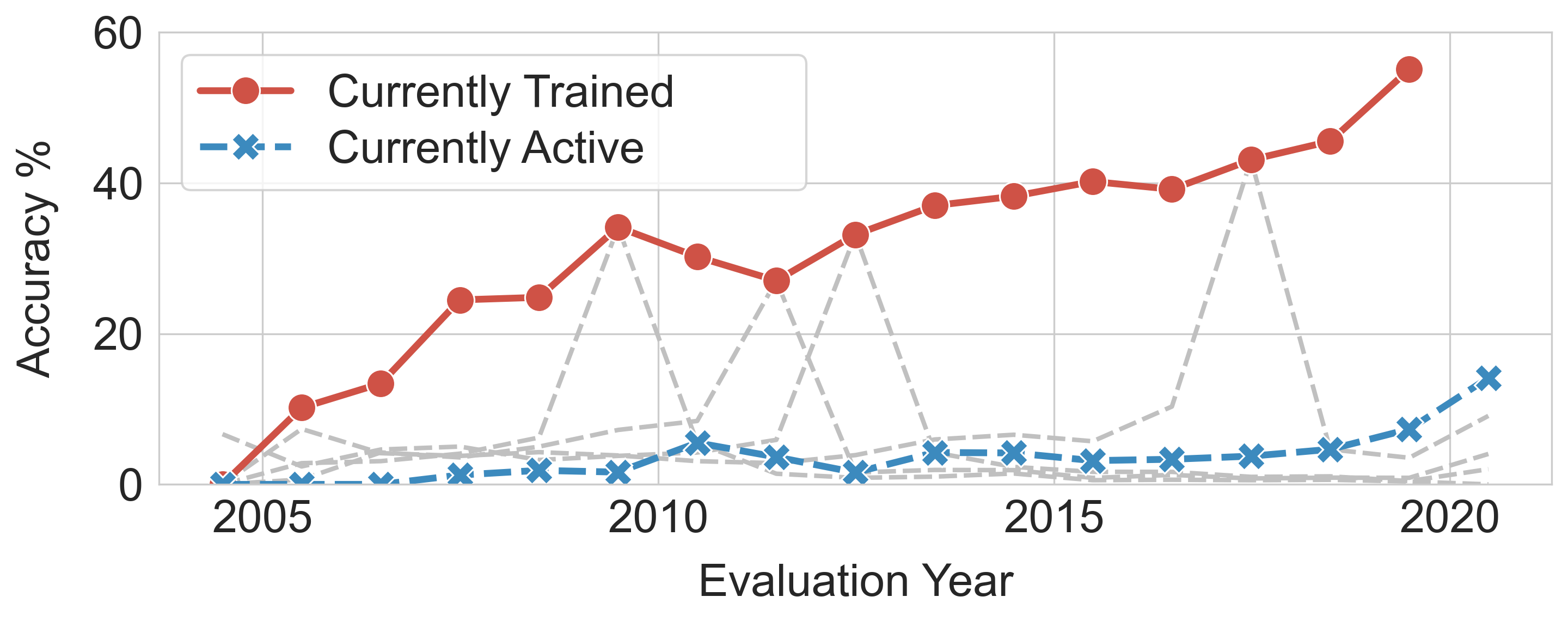}
    \end{adjustbox}
        \caption{Visualization of the currently \textcolor{plotred}{trained} vs. \textcolor{plotblue}{active} composite model on \textsc{CGLM-landmark}. The grey dashed lines are a subset of the models trained during the pipeline.} 
    \label{fig:trained-vs-active}
    \Description{TODO!}
\end{figure}

\textbf{Full data training.} 
This dataset is a good example to showcase the difference between the currently trained and currently active composite model (\Cref{subsec:formal-comparing}).
\revision{As seen in~\Cref{fig:trained-vs-active}, which shows the accuracy sequence \(\Lambda_{\sigma, P}\), the currently trained model has a much higher accuracy over time, since due to its definition it connects the spikes instead of the point after the spike.}
The currently trained numbers are in-line with the numbers by Prabhu et al.~\cite{Prabhu2023CGLM}.
\revision{The reason why the individual models have spikes} is that many classes are mostly prevalent within a single year, i.e., there is a concentration of classes on one particular year.
We explain this with the nature of the dataset: it is likely that landmark pages on Wikipedia get updated in batches, e.g., a user updates pictures of the Big Ben in London in 2015, and then they are not updated for several years again.
Hence, models overfit to the current prevalent classes, forgetting about the old classes.
In a traditional continual learning setup, this might not get noticed.
Full data training has an average top-5 accuracy of 51.5\,\%.

\begin{figure}
    \centering
    \begin{adjustbox}{trim=0cm 0.4cm 0cm 0cm}
                \includesvg[width=0.75\linewidth]{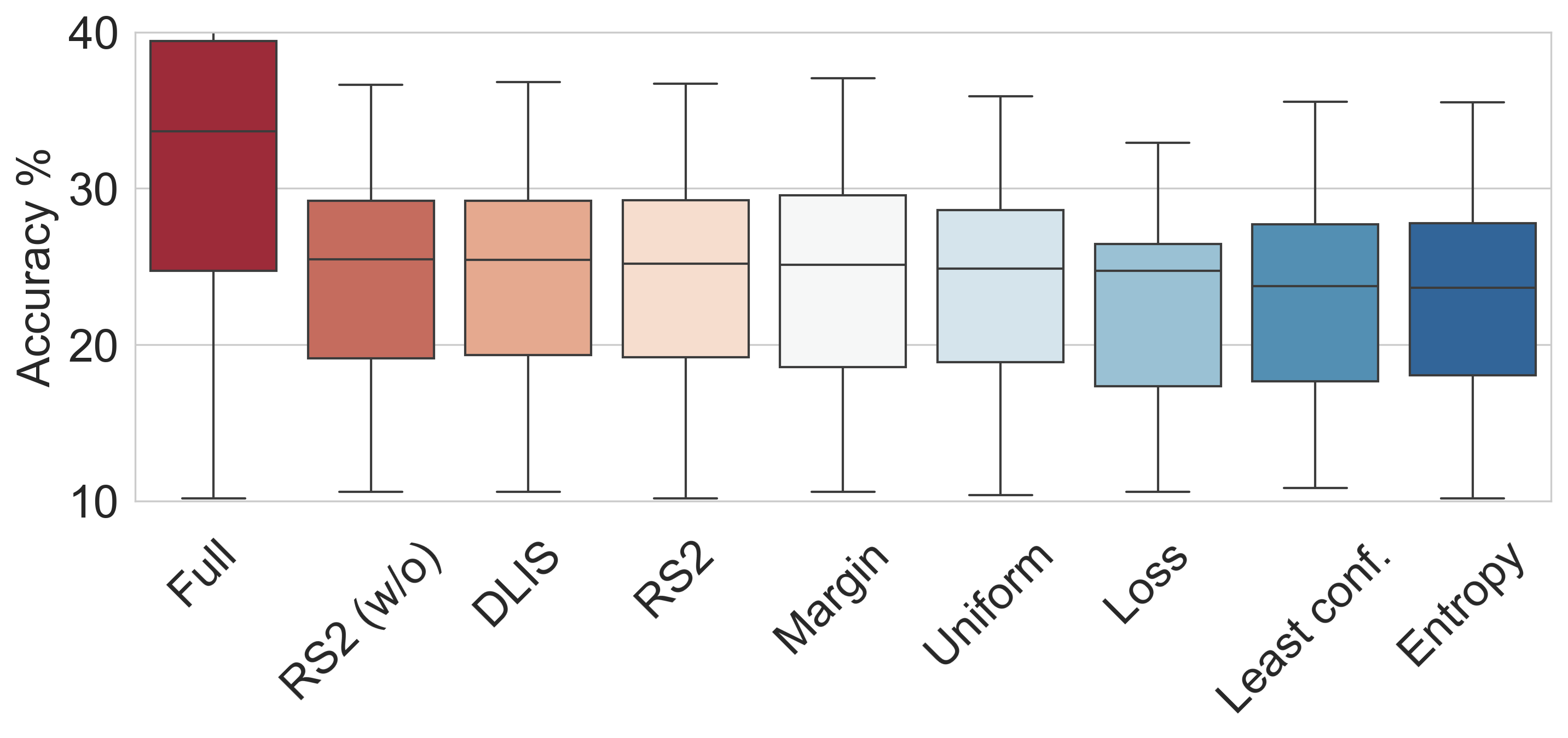}
    \end{adjustbox}
    \caption{Composite model accuracies for full data training and 50\,\% data selection on \textsc{CGLM-landmark}.} 
    \label{fig:cglm-accuracy}
    \Description{TODO!}
    \vspace{-0.5cm}
\end{figure}

\textbf{50\,\%, 25\,\%, and 12.5\,\% subsets.} 
For training on 50\,\% subsets, we show the top-5 accuracies of the composite models for the downsamplers in~\Cref{fig:cglm-accuracy}.
For this dataset with shifts in classes, \textsc{margin} performs best (44\,\%), followed by \textsc{DLIS} (43.1\,\%) and RS2 (43\,\%).
RS2, which simply goes through the dataset as much as possible under the given budget, performs better than more sophisticated techniques like \textsc{least conf.} and \textsc{entropy}.

\revision{On this dataset,} for 25\,\% subsets, \textsc{DLIS} performs best (33.7\,\%), followed by \textsc{RS2} (33.6\,\%).
\textsc{margin} (32.6\,\%) performs worse than \textsc{RS2}.
For 12.5\,\% subsets, \textsc{uniform}, \textsc{RS2}, and \textsc{DLIS} all reach around 23\,\% top-5 accuracy.

\subsubsection{Takeaways}
For \textsc{yearbook}, where we have covariate shift, downsampling helps achieve near full-data performance on a 50\,\% budget.
For \textsc{CGLM-landmark}, where we have prior-probability shift, \textsc{RS2}, \textsc{DLIS} and \textsc{margin} work well.
This is motivating since these cheap sampling strategies do not require subject-specific knowledge.
Future analyses might extend this to information retention~\cite{Prabhu2023CGLM} or more expensive downsamplers like CRAIG~\cite{Mirzasoleiman2020CRAIG}.

\subsection{Triggering Policies}\label{subsec:eval-drift}
We explore triggering policies for full data training on  \textsc{yearbook}, using the setup from~\Cref{subsec:eval-selection}.
\revision{We also explore the Kaggle \textsc{arXiv} dataset to analyze a dataset with a different drift pattern and modality (text).}
We use the number of triggers instead of wall-clock time as a cost metric since we run the experiments on a shared machine.
While all pipelines train on the same number of data points, fewer triggers are desirable due to system overhead per trigger and because the underlying assumption is that we cannot finetune on the fly due to costly deployment checks.

\begin{figure}
    \centering
    \begin{adjustbox}{trim=0cm 0.4cm 0cm 0.5cm}
                \includesvg[width=0.75\linewidth]{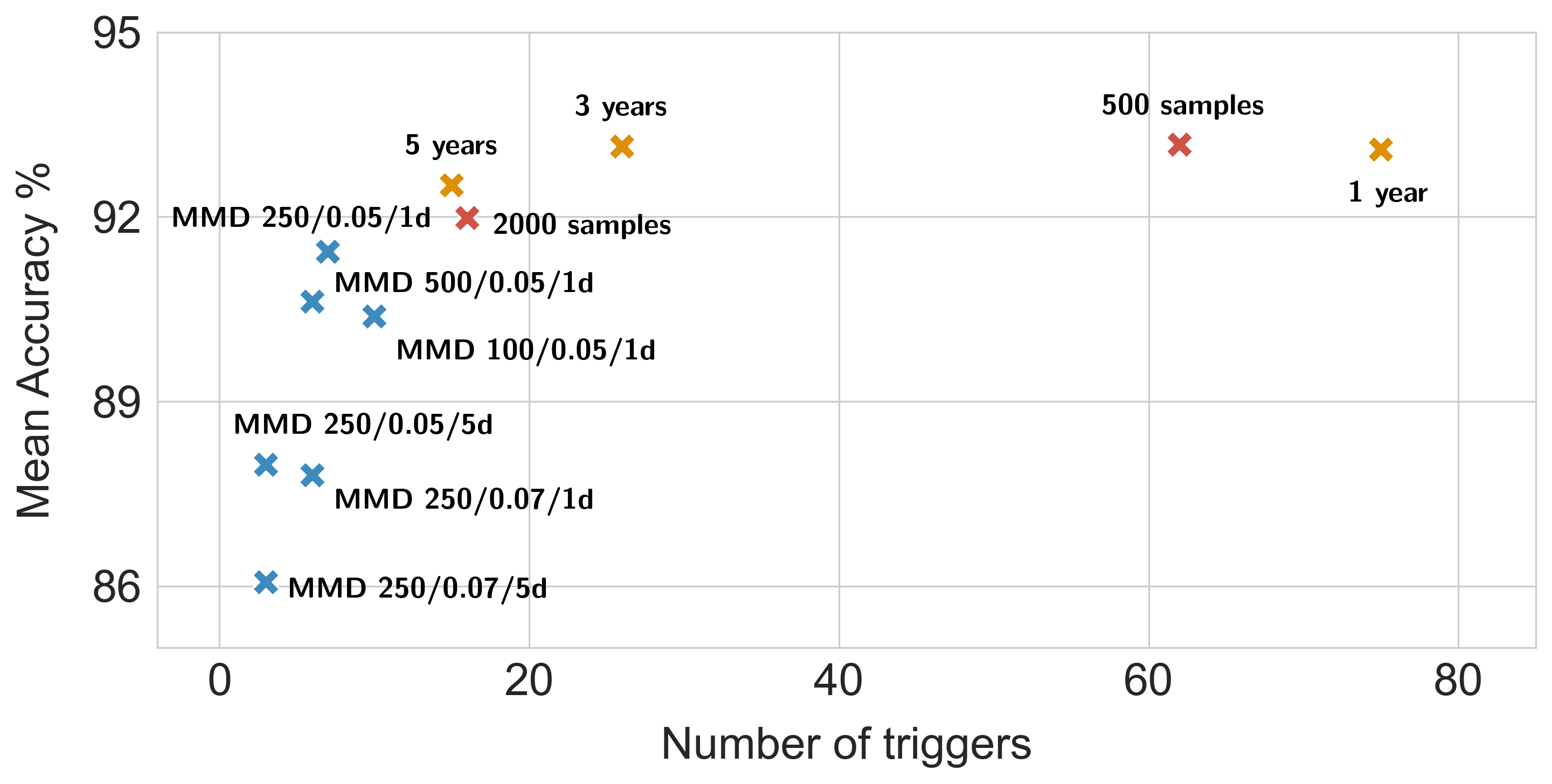}
    \end{adjustbox}
    \caption{\revision{Feasible set of} \revision{triggering polices on \textsc{yearbook}.}} 
    \label{fig:trigger-scatter}
    \Description{TODO!}
    \vspace{-0.5cm}
\end{figure}

\revision{A plot of the cost-accuracy feasible set \(\mathbf{F}\) (\Cref{subsec:formal-comparing}) for different triggering policies is shown in~\Cref{fig:trigger-scatter}.}
We use the currently active model because the currently trained model strongly favors fewer triggers: if we only trigger at the end, the model that has seen all data is by definition the currently trained model for all evaluations and would have very high accuracy.
To fairly compare policies, we only consider the metrics after every pipeline triggered once since there is no active model before the first trigger.
Otherwise, the missing initial values would skew the average.
In general, the goal is to minimize the number of triggers while maximizing accuracy.

\textbf{Time and amount triggers.}
In~\Cref{subsec:eval-selection}, we trigger every year, which is the highest time resolution for \textsc{yearbook}.
Here, we explore triggering every 3 and 5 years, as well as every 500 and 1000 samples. 
Notably, triggering yearly is not optimal: \revision{Triggering every 3 years yields 26 instead of 75 triggers\footnote{As mentioned, we only consider metrics after \emph{all pipelines} triggered once. While the yearly trigger overall fires 84 times, it fires 75 times after all triggers have fired once.}, but only a slightly lower average accuracy (92.8\,\% vs. 93.1\,\%).}
Triggering every 500 items performs similarly due to the even distribution of samples across years.
When we trigger every 5 years, the performance drops to 92.4\,\% accuracy.

\revision{\textbf{Performance triggers.}
These triggers fire when the model performance on a window drops below a threshold.
For the first 3\,500 samples we warm up and trigger at minimum every 3 years.
We use windows of size 250 and test thresholds of 80\,\%, 85\,\%, 90\,\%, and 95\,\% accuracy.
Generally, higher thresholds result in more frequent triggers and improved performance.
Interestingly, the 80\,\% threshold performs better than 85\,\%.
The 80\,\% threshold triggers slightly later, and the resulting model has a better performance than the model from the earlier 85\,\% trigger.
Both models do not cross the threshold for some time, such that the overall average performance of 85\,\% is lower.
\emph{If labels are available}, performance triggers are a simple but well-performing triggering mechanism.
}

\textbf{Drift triggering.}
\revision{The previous triggers rely on prior knowledge: we configure amount and time triggers based on our experience on when drift occurs and how many samples there are.
They also assume a constant drift frequency.
This does not reflect reality where trend seasonality might be irregular~\cite{Mahadevan2023Retraining}.
Performance triggers require labels as well as expected model performance.}
Drift-based policies do not require this prior knowledge, as they use information from the data itself.
\revision{We perform the same warm up as for performance triggers.}
We test MMD (using alibi-detect~\cite{alibi-detect}) on embeddings without PCA, use threshold-based triggering, and sweep across detection intervals (100, 250, 500), thresholds (0.05, 0.07, 0.09), and window sizes (1 day, 5 days) of which we show a subset in~\Cref{fig:trigger-scatter}.
\revision{We also test \modyn's automatic threshold mechanism that triggers when the drift score is in the top 5\,\% of the 15 previously observed scores (\texttt{AutoDrift})}.
\begin{figure}[b]
    \centering
    \begin{adjustbox}{trim=0cm 0.5cm 0cm 0cm}
                \includesvg[width=0.75\linewidth]{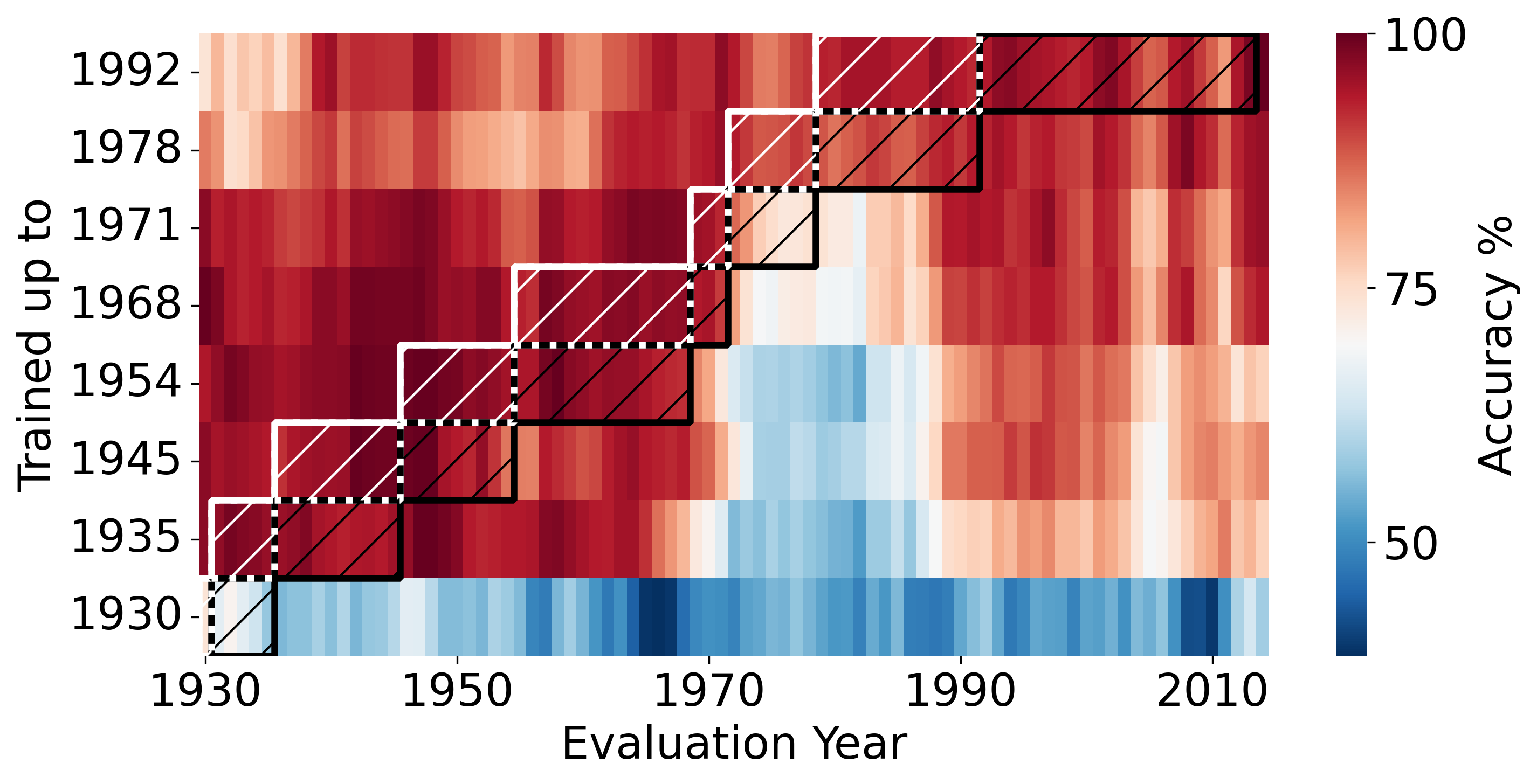}
    \end{adjustbox}
    \caption{The drift MMD (250/0.05/1d) triggering policy on \textsc{yearbook}. \revision{The black boxes indicate when a model is active, i.e., the time during which it would be used for inference.}}
    \label{fig:drift-viz}
    \Description{TODO!}
\end{figure}

\revision{On \textsc{yearbook}}, the drift policies trigger not as often.
\revision{A detection interval of 250 samples, with a threshold of 0.05 and 1 day window performs well, as it only triggers 8 times while still having an average accuracy of 90.1\,\%.}
In~\Cref{fig:drift-viz} we show how the policy navigates around the drift area: consider the model trained up to 1954. 
Shortly before the model's performance degrades in the 1970s, a trigger is fired (end of black box) and the model up to 1968 is trained \revision{(finetuned on the data seen since 1954).}
Note that the drift policy does not have information about future model performance and just uses the information from the data itself to make these decisions.
The other configurations perform slightly worse as they are less sensitive.
For example, increasing the window to 5 days decreases the number of triggers to 3 with an accuracy of 88\,\%.
A larger window size smoothens the drift scores, as the new data needs to be significantly different from the data in the larger window.

\revision{The \texttt{AutoDrift} policy performs well, as it triggers 14 times with an average accuracy of 92.7\,\%.
Importantly, this policy does not require information on the drift metric magnitude.
It uses a simple outlier detection mechanism, making drift detection more user-friendly.}
\revision{Overall, these results are promising as} the drift policies successfully navigate around \revision{\textsc{yearbook}'s} drift area without using prior information on the dataset.

\revision{\textbf{Kaggle \textsc{arXiv} dataset.} The task of this large (\(\sim 2\)\,M samples from 1990 to 2024) textual dataset is to classify paper titles from arXiv into 172 categories.
Textual data uses embeddings for drift detection.
The dataset has a different drift pattern, as performance slowly degrades over time.
We train a DistilBERT model~\cite{Sanh2020Distilbert} with AdamW, learning rate \(0.00002\), and 5 epochs per trigger.
We evaluate each pipeline using 6 month tumbling windows, and warm up the drift and performance triggers for 20\,k samples.}

\begin{figure}
    \centering
    \begin{adjustbox}{trim=0cm 0.4cm 0cm 0.5cm}
                \includesvg[width=0.75\linewidth]{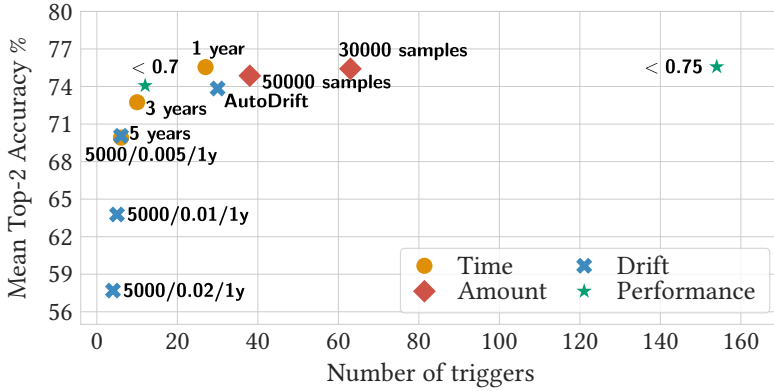}
    \end{adjustbox}
    \caption{\revision{Feasible set of} \revision{triggering polices on Kaggle \textsc{arXiV}.}} 
    \label{fig:trigger-scatter-arxiv}
    \Description{TODO!}
    \vspace{-0.5cm}
\end{figure}

\revision{We show the cost-accuracy scatter in~\Cref{fig:trigger-scatter-arxiv}.
As more papers are submitted each year, the data density increases, and amount triggers fire more frequently than time triggers on this dataset.
Performance triggers strongly depend on the threshold, as the 75\,\% threshold triggers 154 times while 70\,\% triggers only 12 times.
The drift trigger with a threshold of 0.0005 and 1 year windows almost matches the 5 year trigger with 6 triggers and 70\,\% top-2 accuracy.
\texttt{AutoDrift} again performs well with 30 triggers and 73.8\,\% top-2 accuracy, without the need to configure performance- or drift thresholds manually.
Overall, for both \textsc{yearbook} and Kaggle \textsc{arXiv}, data-centric triggering can reduce pipeline cost.}

\subsection{Training Throughput}\label{subsec:eval-tput}

\begin{figure*}
    \centering
    \adjustbox{trim=0cm 0.5cm 0cm 1cm}{%
        \includesvg[width=\textwidth]{img/gridplot_criteo_128.svg}
    }
    \caption{Throughput (x1000) for \textsc{Criteo} (\Cref{subsubsec:eval-criteo-tput}). The first three rows show the results for partitions with 100\,k samples, and the last three rows for partitions with 2.5\,M samples. For each partition, we show results for 1, 2, and 8 threads at storage.} 
    \vspace{-0.5cm}
    \label{fig:criteo-training-throughput}
    \Description{TODO!}
\end{figure*}

In this experiment, we train a model and evaluate the training throughput for different parameters to show how \modyn's optimizations impact training throughput.

\textbf{Setup.} 
We configure the Postgres storage instance to use 96 maximum parallel workers, with 2 maximum workers per gather.
All components are deployed on the same machine, to avoid measuring network bandwidth instead of \modyn{} throughput.
We run all measurements three times and report the average results.

\textbf{Workloads.}
We consider two workloads.
In the first workload, we train a DLRM recommendation model~\cite{Naumov2019DLRM} on the \textsc{Criteo} 1TB click stream dataset~\cite{Criteo2013Dataset}, which provides user data over 24\,days, with roughly 180\,million samples per day.
Given categorical and numerical features, the task is to predict whether a user will click on a suggestion.
We use this scenario because the high number of samples with thousands of samples per file stress-tests \modyn's data-retrieval implementation, in comparison to simpler scenarios such as vision models.
We use NVIDIA's DLRM implementation~\cite{DLRMNvidia} and follow their \enquote{small} setup with a batch size of 65\,536.
At the storage, we use \modyn's \texttt{BinaryFileWrapper}, i.e., the 160\,B samples are stored in a fixed row size binary file format, distributed across files containing ca. 180\,000 samples each.
The bytes parser function at the trainer creates input tensors directly from a \texttt{memoryview} to avoid unnecessary copies.
The second workload trains a ResNet50~\cite{He2016ResNet} on CGLM, as in~\Cref{subsubsec:eval-cglm}.
We use \modyn's \texttt{SingleSampleFileWrapper}, i.e., each sample is stored in one JPEG file.
The bytes parser function converts the data to an RGB \texttt{PIL.Image} on which the dataset applies image augmentations (e.g., resize and crop) to generate a tensor.

\textbf{Throughput measurement.} The size of each partition, as discussed in~\Cref{subsec:impl-selector}, directly dictates the total number of partitions within the trigger training set.
Every worker gets an equal share of each partition.
Note that we do \emph{not} synchronize CUDA after each batch, i.e., we allow PyTorch to perform computation while the next batch is being fetched.
We do not shuffle for this benchmark.
We measure the time from the start of the training loop to the last model update and obtain the throughput by dividing the time by the total number of samples in the trigger.

\subsubsection{\textsc{Criteo} Throughput}\label{subsubsec:eval-criteo-tput}
In~\Cref{fig:criteo-training-throughput}
we show the throughput of training in the \textsc{Criteo} workload.
We test both a partition size of 100\,k (\(\approx 1.53\) batches per partition) and 2.5\,M samples (\(\approx 38.15\) batches per partition).
We first discuss the results for a single thread at the storage, i.e., the top row per partition size.

\textbf{Data loader workers.}
Using one data loader worker and no prefetching, there is no difference between the small and big partitions.
When enabling prefetching of one partition, i.e., loading the next partition into a buffer before its batches are requested, the throughput increases by 1.89x and 1.42x for small and large partitions, respectively.
Note that \emph{prefetching a partition} means that each worker prefetches its share of a partition.
The smaller partitions benefit more from prefetching.

Increasing the number of workers generally increases throughput.
For example, for the large partitions with one prefetched partition, using four workers improves throughput by 3.84x, using eight workers by 7.34x, and using 16 workers by 1.11x, compared to a single worker.
This increase is explained by the ability to fetch the keys and data from selector and storage in parallel, and the parallelization of the bytes-to-tensor transformation.

Notably, in contrast to the single worker scenario, the larger partition size has higher throughput with multiple workers than the smaller partition size.
For example, for 16 workers and with prefetching one partition (16/1/1), the larger partition setting has 2.15x higher throughput than the smaller partition setting.
This is because for the small partitions and 16 workers, a partition does not even cover 10\,\% of a batch.
For larger partitions, the workers have \(\sim 2.5\) batches per partition, which is sufficient to saturate the GPU.
More workers favor larger partition sizes.

\textbf{Additional prefetching.}
We can both prefetch more partitions and request more partitions in parallel.
For the single threaded storage and the smaller partitions, increasing the number of prefetched partitions--while keeping one parallel request--increases throughput, especially for higher number of workers (e.g. 4/6/1, 8/6/1).
However, there are diminishing returns to increasing the number of prefetch partitions.
For example, for four workers, going from 1 (4/1/1) to 2 (4/2/1) prefetched partitions increases throughput by 1.25x, but going from 2 (4/2/1) to 6 (4/6/1) only increases throughput by 1.07x.
As soon as we fill up the buffer faster than data is consumed, there is no benefit from further prefetching data.
When using more workers, the benefit of prefetching more partitions is higher because fixed size partitions are distributed across all workers. 
Prefetching one partition with four workers prefetches the same amount of samples as eight workers that prefetch 2 partitions.

Using more parallel prefetch requests does not improve throughput.
This is explained by the fact that \modyn's components have upper limits of load they can handle: Postgres has a maximum number of worker threads, the number of gRPC worker threads is limited, and the disk holding the databases and dataset has limited bandwidth.
Many parallel requests overload the system.

\textbf{Multi-threaded storage.}
The data retrieval at the storage can use multiple threads (\Cref{subsec:impl-storage}).
We find that using 2 threads increases throughput, but using 8 threads overloads the system and may lead to worse performance.
The throughput increases are higher for smaller number of workers.
For example, for the setting of one worker and no prefetching (1/0/-), on the small partitions, parallelism increases throughput by 1.29x and 1.57x for 2 and 8 threads, respectively.
For 16 workers (16/0/-), increasing the storage threads from 2 to 8 decreases performance to 0.58x.

The reason for the performance decrease with 8 threads is that, while we parallelize data retrieval, there is a limit on the number of parallel Postgres workers.
If 16 workers send a request that gets split upon 8 threads, and each thread emits one query that executes with 2 workers in parallel, we need 256 Postgres workers, amplified with increasing parallel prefetch requests.
Nevertheless, in the following, we show that we reach sufficiently high training throughput.

\begin{figure}[t]
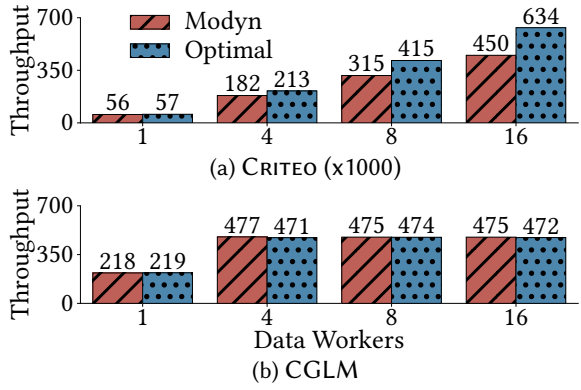

    \centering
    \subfloat[\normalsize \textsc{Criteo} (x1000)]{%
        \adjustbox{trim=0cm 0.15cm 0cm 0.7cm}{%
            \includesvg[width=0.55\linewidth]{img/barplot_local_criteo.svg}
        }
        \label{fig:criteo-local}
    }
    \vspace{-0.45cm}
    \subfloat[\normalsize \textsc{CGLM}]{%
        \adjustbox{trim=0cm 0.15cm 0cm 00cm}{%
            \includesvg[width=0.55\linewidth]{img/barplot_local_cglm.svg}
        }
        \label{fig:cglm-local}
    }
    \vspace{-0.35cm}
    \caption{\textcolor{plotred}{\modyn{} throughput} vs. \textcolor{plotblue}{optimal throughput} when loading data sequentially locally.} 
    \label{fig:local}
    \vspace{-0.5cm}
    \Description{TODO!}
\end{figure}

\textbf{Comparison to local training.}
We compare \modyn{} to local training to quantify its overhead.
For this, we read data sequentially from 90 binary files containing 30\,M samples.
Each dataloader worker is assigned a share of the files.
Note that this not only removes the communication and gRPC overhead, but also removes the sample-level data selection.
\modyn{} loads each sample individually by key, but the local approach loads entire files sequentially and emits all samples in them.

The results are shown in~\Cref{fig:criteo-local}.
For each number of dataloader workers, we compare the best throughput we measure for~\Cref{fig:criteo-training-throughput} against the local throughput.
\modyn{} reaches 98\,\%, 85.4\,\%, 77.8\,\%, and 71\,\% of the optimal local performance for 1, 4, 8, and 16 workers.
Despite having a much more involved data retrieval process, \modyn{} reaches over 70\,\% of optimal throughput for the challenging recommendation system case.

\subsubsection{CGLM Throughput}
\Cref{fig:cglm-local} compares \modyn{} to the optimal local throughput.
As soon as 4 workers are used, the throughput stagnates at around 475\,samples/s.
\modyn{} basically reaches the optimal local throughput for all configurations.
This is because computer vision workloads like CGLM (or \textsc{yearbook}) are \emph{compute-bound}, while training a recommendation systems model is \emph{memory-bound}~\cite{Mudigere2022RecSys, Zhao2022MetaRM,Adnan2021RecSys,Cheng2023DLRMGpu}.
Four workers, with \modyn's C++ storage and selector implementations, supply the model with enough data.

\section{Conclusion and Future Work}\label{sec:conclusion}

We present the data-centric \modyn{} orchestrator for ML pipelines on growing datasets, together with an ecosystem of tooling, benchmarks, and concepts to fairly compare ML pipelines.
\modyn{} implements various triggering and data selection policies and optimizes the system infrastructure under the hood for high-throughput sample-level data selection.
For future work from an ML perspective, it is interesting to extend our analyses across more benchmark datasets, explore more presampling policies, and consider metrics such as information retention~\cite{Cai2021CLOC, Prabhu2023CGLM}.
\revision{Future work might also use \modyn{} and the ideas on comparing pipelines (\Cref{sec:formal-modeling}) to find optimal pipeline configurations on benchmarks with an AutoML approach~\cite{Redyuk2024Pipelines,He2021AutoML}, and extend \modyn{} to the unsupervised case and train generative large language models~\cite{Wang2023LLMDataSurvey}.}
\revision{Due to the right to data deletion in regulations such as GDPR and CCPA~\cite{GDPRArt17,CCPADeletion}, support for data deletion (dynamic instead of just growing datasets) also is an interesting feature~\cite{Bourtoule2021Unlearning,Warnecke2023Unlearning}.}

From a systems and database perspective, additional research opportunities arise.
For example, some selection policies require to store huge embeddings over time~\cite{Pruthi2020Tracin} which is a data management challenge in itself.
It is also not yet clear how to optimally compress and store multiple model versions over time~\cite{Strassenburg2023MMM,Strassenburg2022MMDE}.
\revision{Last, since \modyn{} is a centralized system, it can be leveraged for provenance analyses, such as understanding why retraining and selection decisions were made~\cite{Pina2023Provenance,Chapman2020Provenance,Namaki2020Provenance,Wu2020UpdateProveneance}.}
\modyn{} provides a rich environment for such research on different parts of the training pipeline.

\begin{acks}
\small
Maximilian Böther is supported by the Swiss National Science Foundation (project number 200021\_204620).
Ties Robroek is supported by the Independent Research Fund Denmark’s Sapere Aude program (grant agreement number 0171-00061B).
We thank Francesco Deaglio, Jingyi Zhu, Robin Oester, and Foteini Strati for their contributions to \modyn's codebase.
We also thank the anonymous reviewers for their helpful comments.
\end{acks}

\bibliographystyle{ACM-Reference-Format}
\bibliography{meta/bibliography.bib}

\received{July 2024}
\received[revised]{September 2024}
\received[accepted]{November 2024}

\end{document}